\documentclass[journal]{IEEEtran}
\usepackage{amsmath, amsfonts, amssymb, bm, amsthm}
\usepackage[utf8]{inputenc}
\usepackage{enumitem} 
\usepackage[ruled,vlined]{algorithm2e} 
\makeatletter
\renewcommand\@makefnmark{\hbox{\@textsuperscript{\normalfont\color{black}\@thefnmark}}}
\makeatother

\usepackage{stfloats}
\usepackage[bottom]{footmisc}
\usepackage{color, xcolor, soul, framed}
\usepackage{cancel}
\usepackage[numbers,sort&compress]{natbib}
\usepackage[font=normalfont]{caption} 
\usepackage{diagbox}
\usepackage{pifont}

\usepackage{tikz}
\usetikzlibrary{arrows.meta, positioning, calc, shapes.geometric}

\usepackage{subcaption}
\usepackage{booktabs}
\usepackage{mathtools}
\usepackage{graphicx}
\usepackage{hyperref}
\usepackage{varwidth}
\usepackage{amsmath}
\usepackage{multirow}
\usepackage{array}
\usepackage{lipsum}
\usepackage{orcidlink}

\definecolor{turquoise}{rgb}{.0,.3,1.0}

\usepackage[first=0,last=9]{lcg}

\newcommand{\CC}{\cellcolor{Gray!80!white}}
\usepackage{color, colortbl}
\definecolor{Gray}{gray}{0.93}

\hypersetup{
    colorlinks=true,    citecolor=blue,    linkcolor=red,    filecolor=magenta,  urlcolor=blue,    pdftitle={Overleaf Example},
    }

\begin{document}

\title{C-ZUPT: Stationarity-Aided Aerial Hovering} 
\author{Daniel~Engelsman~\orcidlink{0000-0003-0689-1097},~\IEEEmembership{Graduate Student Member,~IEEE} and~Itzik~Klein~\orcidlink{0000-0001-7846-0654},~\IEEEmembership{Senior Member,~IEEE}
\thanks{The authors are with the Hatter Department of Marine Technologies, Charney School of Marine Sciences, University of Haifa, Israel.\\ E-mails: \{dengelsm@campus, kitzik@univ\}.haifa.ac.il}}


\maketitle
\begin{abstract}
Autonomous systems across diverse domains have underscored the need for drift-resilient state estimation. Although satellite-based positioning and cameras are widely used, they often suffer from limited availability in many environments. As a result, positioning must rely solely on inertial sensors, leading to rapid accuracy degradation over time due to sensor biases and noise. To counteract this, alternative update sources—referred to as information aiding—serve as anchors of certainty. Among these, the zero-velocity update (ZUPT) is particularly effective in providing accurate corrections during stationary intervals, though it is restricted to surface-bound platforms.
This work introduces a controlled ZUPT (C-ZUPT) approach for aerial navigation and control, independent of surface contact. By defining an uncertainty threshold, C-ZUPT identifies quasi-static equilibria to deliver precise velocity updates to the estimation filter. Extensive validation confirms that these opportunistic, high-quality updates significantly reduce inertial drift and control effort. As a result, C-ZUPT mitigates filter divergence and enhances navigation stability, enabling more energy-efficient hovering and substantially extending sustained flight—key advantages for resource-constrained aerial systems.
\end{abstract}
\begin{IEEEkeywords}
Quadrotor dynamics, inertial navigation, optimal LQG control, state estimation, zero-velocity update (ZUPT).
\end{IEEEkeywords}

\section{Introduction}
\IEEEPARstart{D}{espite} substantial progress in navigation technologies, maintaining accurate state estimation over extended durations remains a core challenge—particularly in global navigation satellite systems (GNSS)-denied environments and on highly dynamic platforms. The Kalman filter (KF) is central to most INS/GNSS integrated systems, where it propagates state estimates based on high-frequency inertial measurements, with corrections applied whenever external updates are available. However, disruptions due to signal blockage or communication failures can sever these correction loops, allowing estimation errors to accumulate unchecked due to inertial sensor biases and noise \cite{groves2015principles}.
%
In such situations, any piece of information that enhances situational awareness acts as an island of stability—especially when derived without additional sensors. Aiding mechanisms that exploit motion constraints and environmental interactions, known as information aiding, act as synthetic measurements, providing timely corrections once activation criteria are met \cite{farrell2022inertial, weston2000modern}. 
\\
Among these, Zero-Velocity Updates (ZUPT) have garnered significant attention due to their simplicity, making them one of the most widely adopted techniques in practical navigation systems. Examples include vehicle stops at red lights or in traffic jams \cite{klein2010pseudo}, in-field vehicle calibration \cite{zhuang2015tightly}, pedestrian gait cycles \cite{wang2015stance}, emergency descent in quadrotor failsafe mode \cite{lee2020fail}, surface anchoring \cite{lindve2021tracking}, and underwater docking \cite{yao2017imm}.
\\
Commercial inertial measurement units (IMUs) powered by micro-electromechanical systems (MEMS) technology have become the primary enabler for a wide range of ZUPT-aided applications, extending their use to wearable health devices, pedestrian navigation, autonomous vehicles, indoor robotics, and underwater systems \cite{el2020inertial, cohen2024inertial}.
%
Up until a decade ago, most ZUPT implementations relied on particle filtering \cite{gu2015foot}, analytical methods \cite{wang2018analytical}, or statistical tests \cite{8715398}, with a cost of some precision in favor of improved state estimation. Known for their explainability, simplicity, and ability to model dynamic motion, these methods often used fuzzy logic to handle uncertainty. However, recent machine learning advances have shifted the paradigm, offering greater precision through superior generalization across diverse motion scenarios \cite{wahlstrom2020fifteen}.
\\
Recent literature highlights a broader range of mechanisms beyond ZUPT, including zero-angular rate update (ZARU), zero-down velocity (ZDV), zero-yaw rate (ZYR), constant altitude (CA), and constant position (CP), among others \cite{engelsman2023information}. 
Except in airborne docking contexts \cite{miyazaki2018airborne}, however, these methods have yet to be validated on non-stationary platforms and thus remain limited to terrestrial systems, where ground contact offers a stable reference. Nonetheless, recent advancements in control strategies have enabled aerial platforms to maintain nearly fixed positions while hovering, even in the presence of significant disturbances \cite{tanveer2013stabilized, najm2020altitude}. 
This raises a key question:
\begin{quote}
Can actively controlled equilibrium be leveraged to enable ZUPT during aerial hovering ?
\end{quote}

To answer this, we offer a sensor-free controlled ZUPT (C-ZUPT) state correction approach for aerial navigation without surface contact constraint. Through extensive validation and analysis, the framework centers on three core aspects:
\begin{enumerate}[label=(\roman*)]
    \item \textbf{High-Fidelity Motion Modeling}: An aerial dynamics model supports LQG control under sensor-limited conditions, enabling in-flight disturbance compensation.
    \item \textbf{Drift-Resilient Navigation Framework}: A quadrotor-specific velocity correction method that suppresses drift during stationary phases; publicly available @ \href{https://github.com/ansfl/C-ZUPT}{\texttt{\textbf{GitHub}}}.
    \item \textbf{Battery-Aware Estimation}: An integrated circuit model enables real-time energy monitoring and enforces power constraints, aiding evaluation of the control strategy.
\end{enumerate}
%

\noindent We show that these elements improve estimator–controller interaction, enhance disturbance rejection by 30\%, and reduce long-term energy use—yielding nearly a 6\% efficiency gain. Such enhanced performance can be critical in time-sensitive hovering missions, including urban warfare, policing, rescue, persistent surveillance, infrastructure inspection, and delivery.
\\
The paper proceeds as follows: Section~\ref{sec:theory} covers the theoretical background; Section~\ref{sec:system} describes the system model; Section~\ref{sec:prop} details the proposed approach; Section~\ref{sec:results} analyzes the results; and Section~\ref{sec:conc} concludes the study.

\section{Theoretical Background} \label{sec:theory}
Given the variety of parameters and reference frames in this work, the symbolic framework adopts the following convention: non-bold symbols ($a$) for scalars, bold lowercase ($\boldsymbol{a}$) for vectors, and bold uppercase ($\mathbf{A}$) for matrices. 
\\
The system is modeled as a continuous-time dynamical system, defined by the following differential equation
\begin{align}
\dot{\boldsymbol{x}}(t) = \boldsymbol{f}(\boldsymbol{x}(t), \boldsymbol{u}(t)) \ , \label{eq:sys}
\end{align}
where \( \boldsymbol{x}(t) \in \mathbb{R}^n \) is the state vector, \( \boldsymbol{u}(t) \in \mathbb{R}^k \) is the control input vector, and \( \boldsymbol{f}: \mathbb{R}^n \times \mathbb{R}^k \to \mathbb{R}^n \) is a smooth function that defines the system's dynamics. The solution to this system describes the trajectory of the state \( \boldsymbol{x}(t) \), starting from an initial condition \( \boldsymbol{x}(0) \) and evolving under the influence of the input \( \boldsymbol{u}(t) \). The system is said to be at an equilibrium point, denoted by subscript \( e \), when the dynamics no longer change and the state remains constant, as defined by \cite{arrowsmith1990introduction}
\begin{align}
\dot{\boldsymbol{x}}(t) = \boldsymbol{f}(\boldsymbol{x}_e, \boldsymbol{u}_e) = \boldsymbol{0} \ . \label{eq:sys_eq}
\end{align} 
The behavior of the equilibrium is assessed using the concept of local linearity, which approximates the nonlinear dynamics \( \boldsymbol{f} \) by a linear model near the equilibrium point \( (\boldsymbol{x}_e, \boldsymbol{u}_e) \). The system's response to small perturbations is then studied to classify the equilibrium as stable if trajectories return to it, or unstable if they diverge. 

\begin{table}[b]
\centering
\caption{Characterization of pseudo-measurements (PMs) types: Use-cases, stability, equilibrium, and typical duration.}
\renewcommand{\arraystretch}{1.7}
\begin{tabular}{|c|c|c|c|c|}
\hline
PM & Use-case & Stability & Equilibrium & Time span [sec] \\ \specialrule{1.15pt}{1pt}{1pt} 
ZUPT & \multirow{3}{*}{\shortstack{Vehicular,\\pedestrian,\\indoors.}} & \multirow{3}{*}{Passive} & \multirow{3}{*}{Saddle-type} & \multirow{2}{*}{$10^{-2} \sim 10^{0}$ } \\ \cline{0-0} 
ZARU & & & & \\ \cline{0-0} \cline{5-5}
ZDV & & & & \multirow{2}{*}{$10^{0} \sim 10^{1}$ } \\ \cline{1-4}
ZYR & Track-based & \multirow{3}{*}{Neutral} & \multirow{3}{*}{Stable} & \\ \cline{0-1} \cline{5-5}
CA & \multirow{2}{*}{\shortstack{Calibration\\ \& alignment}} & & & \multirow{2}{*}{$10^{0} \sim 10^{2}$ } \\ \cline{0-0}
CP & & & & \\ \hline 
\end{tabular} \label{t:PMs}
\end{table}

\subsection{Information-aiding}
In navigation systems, dynamic equilibria can be leveraged to improve state estimation accuracy by providing additional constraints. This is achieved through pseudo-measurements (PMs), which are sensor-free observations derived from the platform's assumed physics. By exploiting predictable system behaviors at equilibrium, these equilibria facilitate the indirect estimation of unobservable state variables through the system's dynamics, thereby enhancing both the accuracy and robustness of the state estimation process \cite{farrell2022inertial, weston2000modern, klein2010pseudo}.
\\
Table~\ref{t:PMs} provides a comparative overview of representative aiding mechanisms commonly used in state estimation, categorized by their use case, conditioned stability, and the nature of the exploited equilibrium. As shown in the rightmost column, each constraint is applicable only within a characteristic time scale, determined by how long the system remains sufficiently stationary \cite{engelsman2023information, wang2020review, skog2009car}.
\\
For instance, gait cycles occur frequently but are brief in duration, whereas vehicle stops or platform alignments tend to last longer but occur less frequently. 
\\
Being largely unschedulable, these traditional constraints are ineffective in inherently unstable aerial environments, where near-stationary equilibria can only be achieved through active control.

\subsection{Stable Operating Point}
\begin{figure}[h]
\centering
\begin{tikzpicture}[
  box/.style={draw, thick, minimum width=16mm, minimum height=11mm, fill=gray!6, rounded corners=1mm, inner sep=1.5pt},
  circ/.style={draw, circle, thick, fill=gray!0, minimum size=8.5mm, inner sep=2pt},
  arrow-1/.style={-{Stealth[length=5pt, width=5pt]}, line width=1.pt},
  arrow-2/.style={<->, >={Stealth[length=5pt, width=5pt]}, line width=1.pt}, 
  labelnode/.style={font=\footnotesize, inner sep=1pt},
  every node/.style={font=\footnotesize}
]

\node[circ] (takeoff) {\shortstack{\ Take \vspace{.5mm}\\ \ -off}};
\node[box, right=21mm of takeoff] (hold) {\textbf{Hover}};
\node[circ, right=21mm of hold] (land) {\shortstack{Land \vspace{.5mm}\\ \ -ing}};
\node[box, shift={(-20mm,18mm)}] at (hold) (stabilize) {Stabilize};
\node[box, shift={(+20mm,18mm)}] at (hold) (manual) {\shortstack{Manual \vspace{.5mm}\\ / Acro}};

\node[box, shift={(+20mm,-18mm)}] at (hold) (emergency) {\shortstack{\ Low-Battery \vspace{1mm}\\ / Emergency}};
\node[box, shift={(-20mm,-18mm)}] at (hold) (auto) {\shortstack{\ Autonomous \vspace{1mm}\\ / waypoint}};

\draw[arrow-1] (takeoff) -- (hold);
\draw[arrow-1] (stabilize.east) -- (manual.west);
\draw[arrow-1] (hold) -- (land);
\draw[arrow-2] (stabilize.south) -- (hold.north west);
\draw[arrow-2] (manual.south) -- (hold.north east);
\draw[arrow-1] ([xshift=1mm]manual.south) -- (land.north west);
\draw[arrow-2] (hold.south west) -- (auto.north);
\draw[arrow-2] (hold.south east) -- (emergency.north);
\draw[arrow-1] ([xshift=1mm]emergency.north) -- (land.south west);
\draw[arrow-1] (emergency.west) -- node {\shortstack{Return to Home\vspace{2.5mm}\\ (RTH)}} (auto.east);
\end{tikzpicture}
\caption{State transition diagram showing hovering centrality functioning as a key link among various flight modes \cite{peksa2024review}.}
\label{fig:FMS}
\end{figure}
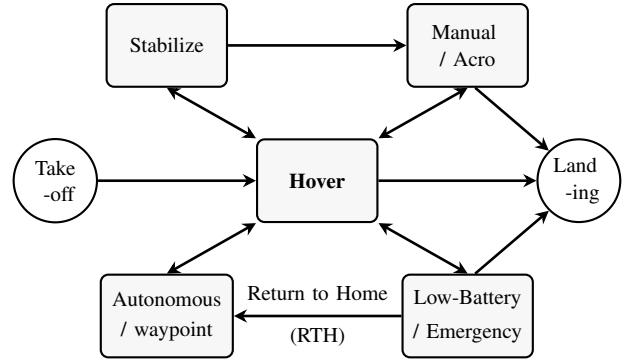

Fixed-wing aircraft typically reach trim conditions during straight and level flight, coordinated turns, and steady climbs or descents. These equilibrium points are linearized to develop an appropriate control scheme, but none of these phases involve zero-velocity timeframes, either translational or angular. In contrast, rotorcraft—especially in hover mode—achieve aerial stationarity by continuously correcting deviations from a fixed position or orientation \cite{saviolo2023learning}.
\\
Recent developments in military, law enforcement, and search and rescue operations have renewed attention on hover mode. While relatively simple, it provides a stable vertical perspective, making it especially effective for tasks requiring precise positioning over extended durations \cite{10530312, bauersfeld2022range, chavez2023learning}.
\\
Flight modes such as takeoff, cruise, and landing are generally time-limited, with hover mode becoming the default when no other commands are issued. This control logic is often represented as a finite state machine, as shown in Fig.~\ref{fig:FMS}, and is commonly implemented in commercial off-the-shelf (COTS) quadrotors, although variations may exist depending on the manufacturer and operating system.
%
\\
Based on linear systems theory, flight conditions where forces and moments are balanced correspond to steady-state dynamics, allowing the nonlinear system \eqref{eq:sys} to be linearized around the equilibrium point \((\boldsymbol{x}_e, \boldsymbol{u}_e)\) through a first-order Taylor expansion \cite{cook2012flight}. Such approximation results in
\begin{align}
\dot{\boldsymbol{x}}(t) \approx \mathbf{A}(\boldsymbol{x}_e, \boldsymbol{u}_e) (\boldsymbol{x}(t) - \boldsymbol{x}_e) + \mathbf{B}(\boldsymbol{x}_e, \boldsymbol{u}_e) (\boldsymbol{u}(t) - \boldsymbol{u}_e) , \label{eq:linearization}
\end{align}
where \( \boldsymbol{x}(t) \) denotes the state vector, \( \boldsymbol{u}(t) \) represents the control inputs, and the deviations from equilibrium are described by
\begin{align}
\delta \boldsymbol{x}(t) = \boldsymbol{x}(t) - \boldsymbol{x}_e \ \ \text{and} \ \ \delta \boldsymbol{u}(t) = \boldsymbol{u}(t) - \boldsymbol{u}_e \ .
\end{align}
Under the assumption of linear time-invariant (LTI) dynamics near the equilibrium points, the Jacobian matrices of the system with respect to the state and input are expressed as
\begin{align}
\mathbf{A}(\boldsymbol{x}_e, \boldsymbol{u}_e) = \frac{\partial \boldsymbol{f}}{\partial \boldsymbol{x}} \bigg|_{(\boldsymbol{x}_e, \boldsymbol{u}_e)} 
\, \text{and} \ \,
\mathbf{B}(\boldsymbol{x}_e, \boldsymbol{u}_e) = \frac{\partial \boldsymbol{f}}{\partial \boldsymbol{u}} \bigg|_{(\boldsymbol{x}_e, \boldsymbol{u}_e)} . \label{eq:Jacobians}
\end{align}
Since $\boldsymbol{x}_e$ and $\boldsymbol{u}_e$ are constants, they vanish upon differentiation. Therefore, the following identities hold
\begin{align}
\boldsymbol{x}(t) = \delta \boldsymbol{x}(t) + \boldsymbol{x}_e \ \ \text{and} \ \ \boldsymbol{u}(t) = \delta \boldsymbol{u}(t) + \boldsymbol{u}_e \ ,
\end{align}
simplifying the error dynamics to the standard LTI form
\begin{align}
\dot{\boldsymbol{x}}(t) &= \mathbf{A} \, \boldsymbol{x}(t) + \mathbf{B} \, \boldsymbol{u}(t) \ , \label{eq:x_gt} \\
\boldsymbol{y}(t) &= \mathbf{C} \, \boldsymbol{x}(t) \ ,
\end{align}
where $\mathbf{C}$ selects the measurable components. 
\\
To incorporate model discrepancies, we introduce zero-mean white Gaussian noise terms, drawn from their respective process (\(\mathbf{W}\)) and measurement (\(\mathbf{V}\)) noise covariance matrices
\begin{align}
\boldsymbol{w}(t) &\sim \mathcal{N}(\mathbf{0}, \mathbf{W}) \ , \quad \mathbb{E}[\boldsymbol{ww}^{\top}] = \mathbf{W} \succ 0 \ , 
\\
\boldsymbol{v}(t) &\sim \mathcal{N}(\mathbf{0}, \mathbf{V}) \ \ , \quad \mathbb{E}[\boldsymbol{vv}^{\top}] = \mathbf{V} \succeq 0 \ . 
\end{align}
These drive the open-loop uncertainty, with covariance matrix ${\mathbf{P}}(t)$ evolve via the differential Lyapunov equation
\begin{align}
\dot{\mathbf{P}}(t) = \mathbf{A} \mathbf{P}(t) + \mathbf{P}(t) \mathbf{A}^\top + \mathbf{W} . \label{eq:state_cov}
\end{align}
Consequently, at each fixed time $t$, the joint distribution of the states and outputs can be approximated as Gaussian
\begin{align}
\begin{bmatrix}
\boldsymbol{x}(t) \\
\boldsymbol{y}(t)
\end{bmatrix}
\hspace{-.5mm} \sim \mathcal{N} \hspace{-.5mm} 
\left(
\begin{bmatrix}
\boldsymbol{\mu}(t) \\
\mathbf{C} \boldsymbol{\mu}(t)
\end{bmatrix} 
, 
\begin{bmatrix}
\mathbf{P}(t) & \mathbf{P}(t) \mathbf{C}^\top \\
\mathbf{C} \mathbf{P}(t) & \mathbf{C} \mathbf{P}(t) \mathbf{C}^\top + \mathbf{V}
\end{bmatrix}
\right) . \label{eq:space}
\end{align}
where $\boldsymbol{\mu}(t)$ denotes the nominal noise-free state trajectory, initialized as $\boldsymbol{\mu} (t_0) = \mathbb{E}[\boldsymbol{x}(t_0)]$, and evolving according to
\begin{align}
\dot{\boldsymbol{\mu}}(t) &= \mathbf{A} \boldsymbol{\mu}(t) + \mathbf{B} \boldsymbol{u}(t) \ . \label{eq:mu_dot}
\end{align}
%

\subsection{Linear–Quadratic–Gaussian Controller} 
\begin{figure}[h] 
\begin{tikzpicture}[ 
  block/.style={draw, thick, minimum width=1.2cm, minimum height=1cm, fill=gray!6, rounded corners=1mm},
  joint/.style={draw, circle, thick, minimum size=6.5mm, inner sep=0pt},
  arrow/.style={-{Stealth[length=6pt, width=6pt]}, rounded corners=1.5mm, line width=1.5pt},
  labelnode/.style={font=\footnotesize, inner sep=1pt},
  every node/.style={font=\footnotesize}
]

\node[joint] (J_sub) {\large $\Delta$};
\node[block, right=9mm of J_sub] (lqr) {LQR \eqref{eq:LQR}};
\node[block, below=10mm of lqr] (kf) {KF \eqref{eq:LQE}};
\node[block, right=10mm of lqr] (plant) {Plant \eqref{eq:x_dot}};
\node[joint, right=10mm of plant] (J_sum) {$\sum$};

\draw[arrow, blue!80!cyan] ([xshift=-10mm]J_sub.west) -- node[xshift=-2mm, above] {$\boldsymbol{x}_{\text{ref}}$} (J_sub);
\draw[arrow, green!75!black] (kf.west) node[xshift=-6.5mm, above] {$\hat{\boldsymbol{x}}$} -| (J_sub.south);
\draw[arrow, green!75!black] (J_sub) -- node[xshift=-1.mm, above] {$\hat{\boldsymbol{\varepsilon}}$} (lqr);
\draw[arrow, blue!80!cyan] (lqr) -- node[xshift=-1.5mm, above] {$\boldsymbol{u}$} (plant);
\draw[arrow, blue!80!cyan] (plant) -- node[xshift=-1.mm, above] {$\mathbf{C} \boldsymbol{x}$} (J_sum);
\draw[arrow, orange!90!black] (J_sum) -- ++(1.1,0) node[xshift=-5.5mm, above] {$\boldsymbol{y}$};
\draw[arrow, blue!80!cyan] (plant) -- (J_sum);

\draw[<-, >={Stealth[length=6pt, width=5pt]}, brown!80!black, line width=1.5pt] (plant.north) -- ++(0,0.55) node[above] {$\boldsymbol{w} \sim \mathcal{N}(\mathbf{0}, \mathbf{W})$};
\draw[<-, >={Stealth[length=6pt, width=5pt]}, brown!80!black, line width=1.5pt] (J_sum.north) -- ++(0,0.73) node[above] {$\boldsymbol{v} \sim \mathcal{N}(\mathbf{0}, \mathbf{V})$};

\draw[arrow, orange!90!black] ([xshift=2mm]J_sum.east) |- ([yshift=-2mm]kf.east);
\draw[arrow, blue!80!cyan, rounded corners=1.5mm] (lqr.east) ++(0.35,0.0) |- ([yshift=2mm]kf.east);

\draw[dashed, line width=1.2pt, rounded corners]($(J_sub.north west)+(-0.45,0.55)$) node[xshift=2cm, above] {LQG} rectangle ($(kf.south east)+(0.72,-0.25)$);
  {\textbf{Your Label}};
\end{tikzpicture}
\caption{Block diagram of a standard LQG closed-loop system. Colored signals indicate: true state (blue), measured output (orange), estimated state (green), and stochastic disturbances (brown). The symbols $\Sigma$ and $\Delta$ represent summation and differencing nodes, respectively.}
\label{fig:LQG}
\end{figure}
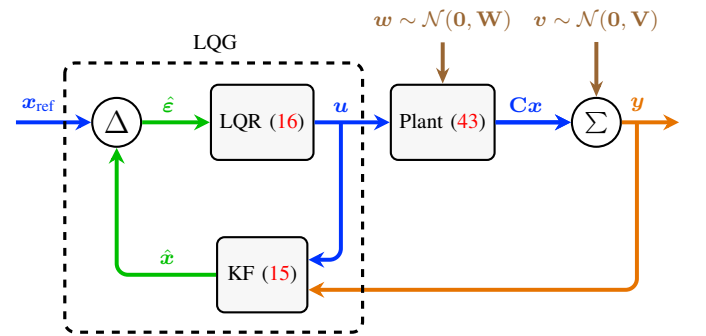

While static feedback assumes full-state access, practical systems typically measure only $m < n$ states. To reconstruct the full state vector from noisy measurements, a linear-quadratic estimator (LQE) is employed \cite{aastrom2012introduction, chrif2014aircraft}. 
\\
For LTI systems, these estimates are then fed into a linear quadratic regulator (LQR), which computes control inputs minimizing a quadratic cost over a finite time horizon $T$, typically defined as 
\begin{align} \label{eq:cost}
\min_{\boldsymbol{u}(t)} J \triangleq \lim_{T \rightarrow \infty} \mathbb{E} \left[  \frac{1}{T} \int_0^T \bigg( \| \boldsymbol{x}(t) \|^2_{\mathbf{Q}} + \| \boldsymbol{u}(t) \|^2_{\mathbf{R}} \bigg) \mathrm{d}t \, \right] ,
\end{align}
subject to \eqref{eq:mu_dot}, with \(\mathbf{Q} \succeq 0\) and \(\mathbf{R} \succ 0\) encoding penalties on state error and input energy, respectively. 
\\
Provided that the pairs \((\mathbf{A}, \mathbf{B})\) and \((\mathbf{A}, \mathbf{W}^{1/2})\) are controllable, and \((\mathbf{C}, \mathbf{A})\) and \((\mathbf{Q}^{1/2}, \mathbf{A})\) are observable, the optimal linear quadratic Gaussian (LQG) controller is obtained by combining the LQR with a Kalman filter, which serves as the LQE. 
Denoting the estimated state by $\hat{\boldsymbol{x}}(t)$, the resulting LQG closed-loop dynamics follow
\begin{align}
\dot{\hat{\boldsymbol{x}}}(t) & = \mathbf{A} \, \hat{\boldsymbol{x}}(t) + \mathbf{B} \, \boldsymbol{u}(t) + \mathbf{L} \left( \boldsymbol{y}(t) - \mathbf{C} \hat{\boldsymbol{x}}(t) \right) \ , \label{eq:LQE} \\
\boldsymbol{u}(t) & = - \mathbf{K} \, \left( \hat{\boldsymbol{x}}(t) - \boldsymbol{x}_{\text{ref}} \right) = - \mathbf{K} \, \hat{\boldsymbol{x}}(t) \ . \label{eq:LQR}
\end{align}
%
The optimal control gain matrix $\mathbf{K}$ is given by
\begin{align}
\mathbf{K} = \mathbf{R}^{-1} \mathbf{B}^\top \mathbf{S} \ ,
\end{align}
where $\mathbf{S}$ is the unique positive semi-definite solution to the algebraic Riccati equation (ARE) given by
\begin{align}
\mathbf{A}^\top \mathbf{S} + \mathbf{S A} - \mathbf{S B R}^{-1} \mathbf{B^\top S + Q} = \boldsymbol{0} \ .
\end{align}
Similarly, the KF optimal gain $\mathbf{L}$ is then given by
\begin{align}
\mathbf{L = P C^\top \left( CPC^\top + V \right)}^{-1} \ ,
\end{align}
where $\mathbf{P}$ is the steady-state estimation error covariance matrix, obtained by solving the following ARE
\begin{align}
\mathbf{A} \mathbf{P} + \mathbf{P} \mathbf{A}^\top - \mathbf{P} \mathbf{C}^\top \mathbf{V}^{-1} \mathbf{C} \mathbf{P} + \mathbf{W} = \mathbf{0} \ .
\end{align}
With \( \boldsymbol{0} \) and \( \mathbf{I} \) representing the zero and identity matrices of the appropriate dimensions, the augmented state space, combining the dynamics from both \eqref{eq:x_gt} and \eqref{eq:LQE}, is defined as follows
\begin{align}
\frac{d}{dt}
\begin{bmatrix}
\boldsymbol{x} \\ \hat{\boldsymbol{x}}
\end{bmatrix} &= \begin{bmatrix}
    \mathbf{A} & \mathbf{-B K} \\ \mathbf{LC} & \mathbf{A-BK-LC}
\end{bmatrix} \begin{bmatrix} \boldsymbol{x} \\ \hat{\boldsymbol{x}} \end{bmatrix} + \begin{bmatrix} \mathbf{I} & \boldsymbol{0} \\ \boldsymbol{0} & \mathbf{L} \end{bmatrix} \begin{bmatrix} \boldsymbol{w} \\ \boldsymbol{v} \end{bmatrix} \, \notag , \\ 
\begin{bmatrix}
\boldsymbol{y} \\ \boldsymbol{u}
\end{bmatrix} &= \begin{bmatrix}
\mathbf{C} & \boldsymbol{0} \\ \boldsymbol{0} & -\mathbf{K}
\end{bmatrix} \begin{bmatrix} \boldsymbol{x} \\ \hat{\boldsymbol{x}} \end{bmatrix} + \begin{bmatrix} \boldsymbol{v} \\ \boldsymbol{0} \end{bmatrix} \ . \label{eq:states_est}
\end{align}
\\
By defining the estimation error as
\begin{align}
\boldsymbol{e}(t) = \boldsymbol{x}(t) - \hat{\boldsymbol{x}}(t) \ , \label{eq:err_est}
\end{align}
and the estimated control error as
\begin{align}
\hat{\boldsymbol{\varepsilon}}(t) = \hat{\boldsymbol{x}}(t) - \boldsymbol{x}_{\text{ref}} \ , \label{eq:err_ctr}
\end{align}
the plant-estimator dynamics admits the block-state form of
\begin{align}
\begin{bmatrix}
    \dot{\boldsymbol{x}} \\ \dot{{\boldsymbol{e}}}
\end{bmatrix} = \begin{bmatrix}
    \mathbf{A-B K} & \mathbf{B K} \\ \boldsymbol{0} & \mathbf{A-LC}
\end{bmatrix} \begin{bmatrix} \boldsymbol{x} \\ {\boldsymbol{e}} \end{bmatrix} + \begin{bmatrix} \mathbf{I} & \boldsymbol{0} \\ \mathbf{I} & -\mathbf{L} \end{bmatrix} \begin{bmatrix} \boldsymbol{w} \\ \boldsymbol{v} \end{bmatrix} \ . \label{eq:error}
\end{align}
This structure highlights the separation principle, with an upper-triangular form that decouples the estimator dynamics (\(\mathbf{A - LC}\)) from the controller dynamics (\(\mathbf{A - BK}\)). 
\\
Fig.~\ref{fig:LQG} illustrates the modular architecture of the LQG framework, in which the KF and LQR components operate independently. This structural independence ensures closed-loop stability without mutual interference \cite{khalil1996robust, mahmudov2000controllability}.

\section{System model} \label{sec:system}
We begin by establishing the assumptions and coordinate frames relevant to hovering, which typically occurs near equilibrium, thereby justifying the standard modeling assumptions common in the literature \cite{hoffmann2007quadrotor, sadr2014dynamics}:
i) The vehicle is a rigid body with symmetric moment of inertia (MoI); ii) Thrust forces and moments are proportional to the square of rotor speeds; iii) Aerodynamic drag is neglected due to low speeds and small cross-sectional area; iv) The center of mass is assumed to coincide with the origin of the body-fixed frame.
%
The orientation of the body frame \(\mathcal{B}\) relative to the inertial frame \(\mathcal{I}\) is given by the rotation matrix  
\begin{align}
\mathcal{R}_{\mathcal{B}}^{\mathcal{I}} := \left( \, \mathbf{e}_x^\mathcal{B}, \, \mathbf{e}_y^\mathcal{B}, \, \mathbf{e}_z^\mathcal{B} \, \right) \in SO(3) \ , \label{eq:frames}
\end{align}  
which is constructed through a sequence of three consecutive rotations: \(\mathcal{R}_z(\psi)\) about the \(\mathbf{e}_z^\mathcal{I}\) axis, followed by \(\mathcal{R}_y(\theta)\) about the \(\mathbf{e}_y^\mathcal{I}\) axis, and \(\mathcal{R}_x(\phi)\) about the \(\mathbf{e}_x^\mathcal{I}\) axis. 
\\
The angles \(\{\phi, \theta\} \in \left( -\frac{\pi}{2}, \frac{\pi}{2}\right)\) and \( \psi \in \left( -\pi, \pi \right)\) correspond to the roll, pitch, and yaw angles, respectively, and define the rotation matrix according to the extrinsic \(z\)-\(y\)-\(x\) convention by 
\begin{equation} \label{eq:T_b_n}
\mathcal{R}_{\mathcal{B}}^{\mathcal{I}} = \begin{pmatrix}
\text{c}_\theta  \text{c}_\psi  &  \text{s}_\phi \text{s}_\theta \text{c}_\psi - \text{c}_\phi \text{s}_\psi  &  \text{c}_\phi \text{s}_\theta \text{c}_\psi + \text{s}_\phi \text{s}_\psi \\
\text{c}_\theta \text{s}_\psi & \text{s}_\phi \text{s}_\theta \text{s}_\psi +  \text{c}_\phi \text{c}_\psi & \text{c}_\phi \text{s}_\theta \text{s}_\psi - \text{s}_\phi \text{c}_\psi  \\
-\text{s}_\theta & \text{s}_\phi \text{c}_\theta & \text{c}_\phi \text{c}_\theta 
\end{pmatrix} \ ,
\end{equation}
\\
where 's' and 'c' represent the sine and cosine functions. Due to the centrality of the vertical axis in both reference frames, the following unit vector will be frequently used
\begin{align}
\textbf{e}^\mathcal{B}_z = (\mathcal{R}_{\mathcal{I}}^{\mathcal{B}}) \, \textbf{e}^\mathcal{I}_z = (\mathcal{R}_{\mathcal{B}}^{\mathcal{I}})^\top 
\begin{pmatrix} 0 \\ 0 \\ 1 \end{pmatrix} = 
\begin{pmatrix}  
- \text{s}_\theta \\
\text{s}_\phi \text{c}_\theta \\
\text{c}_\phi \text{c}_\theta 
\end{pmatrix} \ .
\end{align}
%

\subsection{Airframe Dynamics}
Building on the Newton-Euler formalism \cite{castillo2005modelling}, the coupled translational and rotational dynamics can be expressed as
\begin{align} \label{eq:Newton_Euler}
\begin{pmatrix} 
\boldsymbol{f}^\mathcal{B} \\ \boldsymbol{\tau}^\mathcal{B} 
\end{pmatrix}=
\begin{pmatrix} 
m \boldsymbol{I}_{3} & \boldsymbol{0}_{3} \\ \boldsymbol{0}_{3} & \boldsymbol{J}
\end{pmatrix}
\begin{pmatrix}
\dot{\boldsymbol{v}}^\mathcal{B} \\  \dot{\boldsymbol{\omega}}^\mathcal{B}
\end{pmatrix}
+
\begin{pmatrix} 
\boldsymbol{\omega}^\mathcal{B} \times m \, \boldsymbol{v}^\mathcal{B} \\
\boldsymbol{\omega}^\mathcal{B} \times \boldsymbol{J} \boldsymbol{\omega}^\mathcal{B} 
\end{pmatrix} \, ,
\end{align}
where \( m \) represents the rigid body mass, and \( \boldsymbol{J} \), \( \boldsymbol{I}_3 \), and \( \boldsymbol{0}_3 \) denote the \( 3 \times 3 \) inertia tensor, identity matrix, and zero matrix, respectively. Both the forces $\boldsymbol{f}$ and moments $\boldsymbol{\tau}$ are expressed in the body frame, governing the time derivatives of both the linear velocity \( \boldsymbol{v}^\mathcal{B} \) and the angular velocity \( \boldsymbol{\omega}^\mathcal{B} \).
\\
During fixed-point hovering, near-zero body velocities render drag forces negligible due to their quadratic nature. As a result, the net force in the body frame consists solely of the gravitational force $\textbf{\textit{g}}^\mathcal{B}= \textit{g} \, \textbf{e}^\mathcal{B}_z$, and the vertical rotor thrusts $\boldsymbol{f}_{\textit{z},i}^\mathcal{B}$, resulting in the balance
\begin{align} \label{eq:thrust}
\boldsymbol{f}^\mathcal{B} = \sum_{i=1}^4 \boldsymbol{f}_{\textit{z},i}^\mathcal{B} - m \, \textbf{\textit{g}}^\mathcal{B} \ .
\end{align}
Assuming no blade flapping or variable pitch, each rotor’s spin axis is directed along the body-frame $z$-axis, allowing the thrust generated by the $i$-th propeller to be modeled as
\begin{align} \label{eq:thrust_single}
\boldsymbol{f}^\mathcal{B}_{z,i} = {f}_{z,i} \, \textbf{e}^\mathcal{B}_z = \left( k_T \Omega_i^2 \right) \textbf{e}^\mathcal{B}_z \, ,
\end{align}
where $k_T$ is a lumped coefficient addressed in Sec.~\ref{sub:hover}. 

\begin{figure}[h]
\centering
\begin{tikzpicture}[
arrow/.style={-{Stealth[length=5pt, width=5pt]}, line width=1.75pt},
d_arrow/.style={-{Stealth[length=5pt, width=5pt]}, line width=1.5pt, dashed},
d_line/.style={line width=1.5pt, dashed},
frame/.style={line width=4pt, draw=gray!50},
labelnode/.style={font=\footnotesize, inner sep=1pt},
circ/.style={thick, inner sep=0pt},
]

\node[circ, shift={(+20mm,+15mm)}] at (0,0) (f_1){}; 
\node[circ, shift={(+35mm,-17.5mm)}] at (0,0) (f_2) {};
\node[circ, shift={(-20mm,-15mm)}] at (0,0) (f_3) {};
\node[circ, shift={(-35mm,+17.5mm)}] at (0,0) (f_4) {};
\node[circ, shift={(1.8, -3)}] at (0,0) (cs_I) {};

\draw[frame] (f_1) -- (f_3);
\draw[frame] (f_2) -- (f_4);

\draw[arrow] (0,0) -- (0.8,0.6) node[shift={(2mm,-3mm)}]{$\textbf{e}_x^\mathcal{B}$};
\draw[arrow] (0,0) -- (-0.8,0.4) node[shift={(3mm,2mm)}]{$\textbf{e}_y^\mathcal{B}$};
\draw[arrow] (0,0) -- (0.15,0.9) node[right]{$\textbf{e}_z^\mathcal{B}$};

\draw[arrow, brown!80!black] (f_1.north) -- ++(0.15,0.9) node[above] {$\boldsymbol{f}^{\mathcal{B}}_{z,1}$};
\draw[arrow, blue!60!cyan] ($(f_1)+(-0.3,0)$) arc[start angle=240,end angle=-60,x radius=0.6cm,y radius=0.2cm] node[right=2mm, below] {$\boldsymbol{\Omega}_1$};
\draw[arrow, brown!80!black] (f_2.north) -- ++(0.15,0.9) node[above] {$\boldsymbol{f}^{\mathcal{B}}_{z,2}$};
\draw[arrow, blue!60!cyan] ($(f_2)+(0.3,0)$) arc[start angle=-60,end angle=240,x radius=0.6cm,y radius=0.2cm] node[right=5mm, below] {$\boldsymbol{\Omega}_2$};
\draw[arrow, brown!80!black] (f_3.north) -- ++(0.15,0.9) node[above] {$\boldsymbol{f}^{\mathcal{B}}_{z,3}$};
\draw[arrow, blue!60!cyan] ($(f_3)+(-0.3,0)$) arc[start angle=240,end angle=-60,x radius=0.6cm,y radius=0.2cm] node[left=5mm, below] {$\boldsymbol{\Omega}_3$};
\draw[arrow, brown!80!black] (f_4.north) -- ++(0.15,0.9) node[above] {$\boldsymbol{f}^{\mathcal{B}}_{z,4}$};
\draw[arrow, blue!60!cyan] ($(f_4)+(0.3,0)$) arc[start angle=-60,end angle=240,x radius=0.6cm,y radius=0.2cm] node[left, below] {$\boldsymbol{\Omega}_4$};

\draw[arrow, green!70!black] (0.1666,1.0) -- ++(0.175*1.4,1.05*1.4) node[above] {$\boldsymbol{u}_{z}$};
\draw[arrow, green!70!black] ($(0.6,1.5)$) arc[start angle=-60,end angle=240,x radius=0.6cm,y radius=0.2cm] node[above=1mm, left=2mm] {$\boldsymbol{u}_{\psi}$};
\draw[arrow, green!70!black] ($(-2.3,0.6)$) arc[start angle=200,end angle=40,x radius=0.45cm,y radius=0.55cm] node[shift={(2.5mm,-1mm)}] {$\boldsymbol{u}_{\theta}$};
\draw[arrow, green!70!black] ($(0.9,1.)$) arc[start angle=160,end angle=-10,x radius=0.5cm,y radius=0.4cm] node[right=2mm, below] {$\boldsymbol{u}_{\phi}$};

\draw[d_line] (0,0) -- ++(1.8, -3) node[shift={(-5mm,15.5mm)}]{$\boldsymbol{\xi}^\mathcal{I}$};
\draw[d_arrow] (0,0) -- (0,-11mm) node[below]{$m \, \textbf{\textit{g}}^\mathcal{I}$};
\draw[arrow] (cs_I) -- (2.9,-3.1) node[shift={(2mm,-3mm)}]{$\textbf{e}_x^\mathcal{I}$};
\draw[arrow] (cs_I) -- (2.7,-2.6) node[shift={(3mm,2mm)}]{$\textbf{e}_y^\mathcal{I}$};
\draw[arrow] (cs_I) -- (1.85,-2.) node[right]{$\textbf{e}_z^\mathcal{I}$};
\end{tikzpicture}
\caption{Free-body diagram of the quadrotor system model.}
\label{fig:Quad_model}
\end{figure}
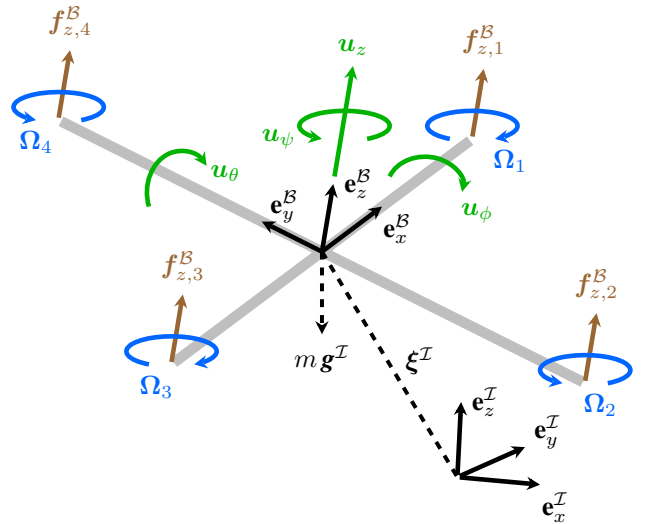

Attitude stabilization relies on differential thrust between two opposing rotors located at a distance $\ell$ from the center of mass. The rolling moment is generated by adjusting the lateral thrust difference between motors \#4 (left) and \#2 (right), given by
\begin{align}
\boldsymbol{\tau}_{\phi}^\mathcal{B} = \ell \left( \boldsymbol{f}^\mathcal{B}_{z,4} - \boldsymbol{f}^\mathcal{B}_{z,2} \right) \textbf{e}^\mathcal{B}_x \, ,
\end{align}
while the pitch moment is controlled by the longitudinal thrust difference between motors \#1 (front) and \#3 (rear), as
\begin{align}
\boldsymbol{\tau}_{\theta}^\mathcal{B} = \ell \left( \boldsymbol{f}^\mathcal{B}_{z,3} - \boldsymbol{f}^\mathcal{B}_{z,1} \right) \textbf{e}^\mathcal{B}_y \, .
\end{align}
Lastly, yawing occurs when the individual rotor drag torques \( \tau_{M,i} \) sum to a nonzero value, reflecting an asymmetry. With \( k_M \) denoting the torque coefficient and positive yaw defined as counter-clockwise about \( \mathbf{e}_z^\mathcal{B} \), the resulting yaw torque is
\begin{align} \label{eq:drag_torque}
\boldsymbol{\tau}_{\psi}^\mathcal{B} = \sum_{i=1}^4 \tau_{M,i} = (-1)^i \left( k_M \Omega_i^2 \right) \textbf{e}^\mathcal{B}_z \, .
\end{align}
%
Accounting for these propulsive effects, the six-degree-of-freedom (6-DoF) motion model in \eqref{eq:Newton_Euler} can be derived as
\begin{align}
\begin{pmatrix}
\dot{\boldsymbol{v}}^\mathcal{B} \\  \dot{\boldsymbol{\omega}}^\mathcal{B} 
\end{pmatrix} = 
\begin{pmatrix}
m^{-1} \boldsymbol{f}^\mathcal{B} - {\boldsymbol{\omega}}^\mathcal{B} \times {\boldsymbol{v}}^\mathcal{B} - \textbf{\textit{g}}^\mathcal{B} \\ 
\boldsymbol{J}^{-1} \left( \boldsymbol{\tau}^\mathcal{B} - \boldsymbol{\omega}^\mathcal{B} \times \boldsymbol{J} \boldsymbol{\omega}^\mathcal{B} \right)
\end{pmatrix} , \label{eq:newton_euler_1}
\end{align}
where the thrust and torques are generated by the rotor speeds according to the control allocation (mixer) matrix
\begin{align}
\begin{pmatrix}
{f}_{z} \\ 
{\tau}_{\phi} \\  
{\tau}_{\theta} \\  
{\tau}_{\psi}
\end{pmatrix} = \begin{pmatrix}
k_T & k_T & k_T & k_T \\ 
0 & - \ell k_T & 0 & \ell k_T \\ 
- \ell k_T & 0 & \ell k_T & 0 \\ 
-k_M & k_M & -k_M & k_M
\end{pmatrix} \begin{pmatrix}
\Omega_1^2 \\ \Omega_2^2 \\ \Omega_3^2 \\ \Omega_4^2
\end{pmatrix} , \label{eq:mixer_scalar}
\end{align}
or, in a more compact form using Hadamard power
\begin{align} \label{eq:mixer_matrix}
\boldsymbol{u} = \mathbf{M} \, \boldsymbol{\Omega}^{\circ 2} \, .
\end{align}
%
Fig.~\ref{fig:Quad_model} summarizes the key components of the system model, including the inertial frame \(\mathcal{I} = (\mathbf{e}_x^\mathcal{I}, \mathbf{e}_y^\mathcal{I}, \mathbf{e}_z^\mathcal{I})\), the body-fixed frame \(\mathcal{B} = (\mathbf{e}_x^\mathcal{B}, \mathbf{e}_y^\mathcal{B}, \mathbf{e}_z^\mathcal{B})\), the quadrotor structure (thick gray), reference axes (black), rotor thrust vectors (brown), control inputs (green), and motor speed actuation (blue).
%

\newpage
\subsection{Kinematics}
The platform's trajectory is derived by time-integrating its dynamics and applying a coordinate transformation to an inertial reference frame, letting the position vector be \cite{etkin1995dynamics}
\begin{align}
{\boldsymbol{\xi}}^\mathcal{I} = \left(x, y, z \right)^\top \, ,
\end{align}
with the corresponding body velocities related by
\begin{align}
\dot{\boldsymbol{\xi}}^\mathcal{I} = \mathcal{R}_{\mathcal{B}}^{\mathcal{I}} {\boldsymbol{v}}^\mathcal{B} \, .
\end{align}
The Euler angles vector in the inertial frame is denoted as
\begin{align}
{\boldsymbol{\eta}}^\mathcal{I} = (\phi, \theta, \psi)^\top \, ,
\end{align}
with its rate of change related to the body angular rates \( (p, q, r) \) through the transformation matrix \( \textbf{\textit{W}}_{\boldsymbol{\eta}} \), such that
\begin{align}
\dot{\boldsymbol{\eta}}^\mathcal{I} = \textbf{\textit{W}}_{\boldsymbol{\eta}} \, {\boldsymbol{\omega}}^\mathcal{B} = 
\begin{pmatrix} 
1 & \text{s}_\phi \text{t}_\theta & \text{c}_\phi \text{t}_\theta \\
0 &  \text{c}_\phi & -\text{s}_\phi \\
0 & \text{s}_\phi / \text{c}_\theta & \text{c}_\phi / \text{c}_\theta 
\end{pmatrix} 
\begin{pmatrix} p \\ q \\ r \end{pmatrix} \, , \label{eq:W_eta}
\end{align}
where ’t’ is the tangent function. For derivation, thrust and torques are idealized as direct control input functions, hence
\begin{align} \label{eq:control}
\boldsymbol{u} (\Omega) = 
\begin{pmatrix}
\boldsymbol{u}_z \\ \boldsymbol{u}_\phi \\ \boldsymbol{u}_\theta \\ \boldsymbol{u}_\psi 
\end{pmatrix} 
= 
\begin{pmatrix}
{f}_{z} \, \textbf{e}_z^\mathcal{B} \\ 
{\tau}_{\phi} \, \textbf{e}_x^\mathcal{B} \\  
{\tau}_{\theta} \, \textbf{e}_y^\mathcal{B} \\  
{\tau}_{\psi} \, \textbf{e}_z^\mathcal{B}
\end{pmatrix} \in \ \mathbb{R}^4 \ .
\end{align}
Finally, by incorporating the generalized coordinates into the system model, the augmented state vector is defined as
\begin{align}
\boldsymbol{x} = \begin{pmatrix} \,
({\boldsymbol{\xi}}^\mathcal{I})^\top & ({\boldsymbol{v}}^\mathcal{B})^\top & ({\boldsymbol{\eta}}^\mathcal{I})^\top & ({\boldsymbol{\omega}}^\mathcal{B})^\top \
\end{pmatrix}^\top \, \in \mathbb{R}^{12} \, , \label{eq:state_vector}
\end{align}
with the nonlinear state-space model in \eqref{eq:sys} evolves as
\begin{align}
\begin{pmatrix}
\dot{\boldsymbol{\xi}}^\mathcal{I} \\ \dot{\boldsymbol{v}}^\mathcal{B} \\ \dot{\boldsymbol{\eta}}^\mathcal{I} \\ \dot{\boldsymbol{\omega}}^\mathcal{B} 
\end{pmatrix} = 
\begin{pmatrix}
(\mathcal{R}_{\mathcal{B}}^{\mathcal{I}}) \, {\boldsymbol{v}}^\mathcal{B} \\
m^{-1} \boldsymbol{u}_z - {\boldsymbol{\omega}}^\mathcal{B} \times {\boldsymbol{v}}^\mathcal{B} - {\textit{g}} \, \textbf{e}^\mathcal{B}_z \\ 
(\textbf{\textit{W}}_{\boldsymbol{\eta}}) \boldsymbol{\omega}^\mathcal{B} \\ 
\boldsymbol{J}^{-1} \left( ( \boldsymbol{u}_\phi + \boldsymbol{u}_\theta + \boldsymbol{u}_\psi ) - \boldsymbol{\omega}^\mathcal{B} \times \boldsymbol{J} \boldsymbol{\omega}^\mathcal{B} \right)
\end{pmatrix} , \label{eq:x_dot}
\end{align}
with a step-by-step derivation provided in Appendix~\ref{appendix:a}.

\section{Proposed Approach} \label{sec:prop}
We begin by analyzing the near-stationarity dynamics at a local hovering equilibrium, followed by an observability analysis that motivates the proposed aiding mechanism.

\subsection{Motivation}
Since hovering occurs near equilibrium with small state variables, local behavior can be approximated using small-angle assumptions. This simplifies \eqref{eq:W_eta} and \eqref{eq:T_b_n} to $\dot{\boldsymbol{\eta}}^\mathcal{I} \approx \boldsymbol{\omega}^\mathcal{B}$ and $\mathcal{R}_{\mathcal{B}}^{\mathcal{I}} \approx \boldsymbol{I}_3 + (\boldsymbol{\eta})_{\times}$, respectively, where $(\cdot)_{\times}$ denoting the skew-symmetric operator. With these approximations, the Jacobian of the nonlinear dynamics in equation \eqref{eq:x_dot} is computed as
\begin{align}
\frac{\partial \boldsymbol{f}}{\partial \boldsymbol{x}} = 
\begin{pmatrix}
\boldsymbol{0}_3 & \boldsymbol{I}_{3} + (\boldsymbol{\eta})_\times & -({\boldsymbol{v}}^\mathcal{B})_\times & \boldsymbol{0}_3 \\ 
\boldsymbol{0}_3 & -({\boldsymbol{\omega}}^\mathcal{B})_\times & - (\textbf{\textit{g}}^\mathcal{I})_\times & ({\boldsymbol{v}}^\mathcal{B})_\times \\ 
\boldsymbol{0}_3 & \boldsymbol{0}_3 & \boldsymbol{0}_3 & \boldsymbol{I}_3 \\ 
\boldsymbol{0}_3 & \boldsymbol{0}_3 & \boldsymbol{0}_3 & \nabla_{\boldsymbol{\omega}} (\dot{\boldsymbol{\omega}}^\mathcal{B} )
\end{pmatrix} ,
\end{align}
with the corresponding sub-Jacobian of $\dot{\boldsymbol{\omega}}^\mathcal{B}$ given by
\begin{align}
\nabla_{\boldsymbol{\omega}} (\dot{\boldsymbol{\omega}}^\mathcal{B} ) = \boldsymbol{J}^{-1} \left( \, ( \boldsymbol{J} \boldsymbol{\omega}^\mathcal{B} )_{\times} - \boldsymbol{\omega}^\mathcal{B} \times \boldsymbol{J} \, \right) \, .
\end{align}
Assuming a hovering equilibrium defined by a state vector at a fixed point and a nominal thrust control input
\begin{align}
\boldsymbol{x}_e &= \begin{pmatrix} \boldsymbol{\xi}^\mathcal{I}_e & \textbf{0}_3 & {\boldsymbol{\eta}}^\mathcal{I}_e & \textbf{0}_3 \end{pmatrix}^\top \ \in \ \mathbb{R}^{12} \ , \label{eq:equilibrium_state} \\
\boldsymbol{u}_e &= \begin{pmatrix} m \textit{g} & 0 & 0 & 0 \end{pmatrix}^\top \ \in \ \mathbb{R}^{4} \ , \label{eq:equilibrium_input}
\end{align}
substituting this operating point into the state Jacobian yields the linearized system dynamics model
\begin{align} \label{eq:A_mat}
\mathbf{A} = \frac{\partial \boldsymbol{f}}{\partial \boldsymbol{x}}  \bigg|_{(\boldsymbol{x}_e, \boldsymbol{u}_e)} = \hspace{5cm} \\ \notag
\begin{pmatrix}
\boldsymbol{0}_3 & \boldsymbol{I}_3 + (\boldsymbol{\eta}_e)_\times & \boldsymbol{0}_3 & \boldsymbol{0}_3 \\ 
\boldsymbol{0}_3 & \boldsymbol{0}_3 & - (\textbf{\textit{g}}^\mathcal{I})_\times & \boldsymbol{0}_3 \\ 
\boldsymbol{0}_3 & \boldsymbol{0}_3 & \boldsymbol{0}_3 & \boldsymbol{I}_3 \\ 
\boldsymbol{0}_3 & \boldsymbol{0}_3 & \boldsymbol{0}_3 & \boldsymbol{0}_3 
\end{pmatrix} \in \, \mathbb{R}^{12 \times 12} ,
\end{align}
\\
where higher-order centrifugal terms and their Coriolis couplings vanish. Likewise, linearizing the control input around the thrust required to balance gravity gives
\begin{align}
\mathbf{B} = \frac{\partial \boldsymbol{f}}{\partial \boldsymbol{u}} \bigg|_{(\boldsymbol{x}_e, \boldsymbol{u}_e)} = \begin{pmatrix} 
\boldsymbol{0}_{3 \times 1} & \boldsymbol{0}_{3 \times 3} \\ 
m^{-1} \textbf{e}_z^{\mathcal{B}} & \boldsymbol{0}_{3 \times 3} \\ 
\boldsymbol{0}_{3 \times 1} & \boldsymbol{0}_{3 \times 3} \\ 
\boldsymbol{0}_{3 \times 1} & \boldsymbol{J}^{-1}
\end{pmatrix} \in \, \mathbb{R}^{12 \times 4} \, .
\end{align}
Within the state estimation process, these linearized dynamics are propagated forward using three key measurable quantities:
\\
\subsubsection{Inertial Measurements} High-rate IMU data propagate the state via the process model but, without external or information-based aiding, offer no direct observability, leading to unbounded estimation uncertainty.
\\
\subsubsection{Position Measurements} Provides corrective feedback to all states except yaw ($\psi$), but is inherently limited by satellite access. The associated measurement model is given by
\begin{align} \label{eq:obs_1}
\mathbf{C}_{\boldsymbol{\xi}} = 
\left(
\begin{array}{cccc}
\boldsymbol{I}_3 & \boldsymbol{0}_3 & \boldsymbol{0}_3 & \boldsymbol{0}_3
\end{array}
\right) .
\end{align}
\subsubsection{Velocity Measurements} whether instrumented or ZUPT-based, constrain velocity errors and related states, mapped as
\begin{align} \label{eq:obs_2}
\mathbf{C}_{\boldsymbol{v}} = 
\left(
\begin{array}{cccc}
\boldsymbol{0}_3 & \boldsymbol{I}_3 & \boldsymbol{0}_3 & \boldsymbol{0}_3
\end{array}
\right) .
\end{align}
Therefore, applying the LTI observability criterion \cite{ramanandan2011observability}
\begin{align} 
\mathcal{O} = \begin{pmatrix} \label{eq:obs_3}
\, \mathbf{C}^\top & (\mathbf{C} \mathbf{A})^\top & \dots & (\mathbf{C} \mathbf{A}^{n-1})^\top
\end{pmatrix}^\top ,
\end{align}
to the system matrices \eqref{eq:A_mat}-\eqref{eq:obs_3} reveals the observability rank of each modality, as evaluated and compared in Table~\ref{t:observ_1}.
\begin{table}[h]
\centering
\caption{Observability ranking across measurement types.}
\renewcommand{\arraystretch}{1.35}
\begin{tabular}{|c|c|c|c|c|c|} \hline
\CC Term & Data\CC & Source\CC& Phase\CC & $\operatorname{rank} (\mathcal{O})$\CC & Sensor-free? \CC \\ \hline
$\mathbf{B}$ & Inertial & IMU & Prediction & 0$/$12 & \ding{55} \\ \hline \hline
$\mathbf{C}_{\boldsymbol{\xi}}$ & Position & GNSS & \multirow{2}{*}{Update} & 10$/$12 & \ding{55} \\ \cline{0-2} \cline{5-6}
$\mathbf{C}_{\boldsymbol{v}}$ & \textbf{Velocity} & \textbf{C-ZUPT} & & \textbf{7}$/$12 & \ding{51}  \\  \hline
\end{tabular} \label{t:observ_1}
\end{table}
\\
As observed, C-ZUPT attains a rank of 7 during GNSS outages—entirely sensor-free. In summary, the linearized dynamics about the hover equilibrium $(\boldsymbol{x}_e, \boldsymbol{u}_e)$ propagate state estimates via the following state-space model
\begin{align}
\hat{\boldsymbol{x}}_{k} &= \mathbf{A} \, \hat{\boldsymbol{x}}_{k-1} + \mathbf{B} \, \boldsymbol{u}_{k-1} \ , \label{eq:prediction} \\ 
\boldsymbol{y}_{k} &= \mathbf{C} \, \boldsymbol{x}_{k} \ .
\end{align}

\subsection{Stationarity Heuristics} \label{sec:zupt}
While numerous ZUPT methods have been developed for surface-bound platforms \cite{el2020inertial, cohen2024inertial, gu2015foot, wang2018analytical, 8715398}, aerial systems rarely maintain true zero-velocity during hover. Consequently, detecting near-stationary conditions involves a trade-off, wherein some precision is deliberately sacrificed to enable timely updates.
\\ 
Algorithm~\ref{alg:zupt} outlines our stationarity detection heuristic, which computes the L2 norms of the estimated body velocities, $\hat{\boldsymbol{v}}^\mathcal{B}$, and specific forces, $\tilde{\boldsymbol{f}}$, averaged over a sliding window spanning the past \( K \) samples. 
Once these norms fall below the predefined thresholds \( \delta_{\tilde{f}} \) and \( \delta_{\hat{v}} \), a loop closure is triggered, and ZUPT assigns the information aiding as
\begin{align}
\boldsymbol{y}_k = \mathbf{C}_{\boldsymbol{v}} \, \boldsymbol{x}_k =
\boldsymbol{v}^{\mathcal{B}}_0 = \boldsymbol{0}_3 \, .\label{eq:v_zupt}
\end{align}
Otherwise, the system proceeds with state prediction. 
\begin{algorithm}[h]
\caption{Stationarity detection and ZUPT generation.}
\SetAlgoLined
\SetKwInOut{Input}{Input}
\Input{$\tilde{\boldsymbol{f}} \, , \, \boldsymbol{x}_k \, , \, \delta_f \, , \, \delta_{\hat{v}} \, , \, K \, .$
}
\BlankLine
\While{\textsf{filter is running}}{
\vspace{1mm} \eIf{ $ \| \tilde{\boldsymbol{f}}_{_{[ k-K:k ]}} \|_2 < \delta_f$  \textsf{and} $ \| \hat{\boldsymbol{v}}^\mathcal{B}_{_{[k-K:k ]}} \|_2 < \delta_{\hat{v}}$}{ 
    $\boldsymbol{v}^\mathcal{B}_{_{ZUPT}} = \boldsymbol{0}$ \\
        $\boldsymbol{y}_k = \mathbf{C}_{\boldsymbol{v}} \, \boldsymbol{x}_k = \boldsymbol{0}_3$ \\
        \KwRet{\big($\hat{\boldsymbol{x}}_{k|k}, \mathbf{P}_{k|k}$\big)} \hspace{13mm} \tcp{Update}
    }
    {
        \KwRet{\big($\hat{\boldsymbol{x}}_{k|k-1}, \mathbf{P}_{k|k-1}$\big)} \hspace{6mm} \tcp{Prediction}
    }   
}
\label{alg:zupt}
\end{algorithm}

\subsection{Control Signal Flow} 
While the idealized dynamics assume direct application of the LQR control input, real-world systems introduce delays and nonlinearities that alter the actual output $\boldsymbol{u}_{\text{out}}$. To reflect these effects in the process model, we identify three key transitional stages that significantly influence the signal flow.
\\
\subsubsection{Command Processing} The LQR control law onboard the flight controller \eqref{eq:LQR} computes the net forces and moments to minimize tracking error \cite{chrif2014aircraft}. 
However, due to the structure of the mixer matrix \eqref{eq:mixer_matrix}, directly inverting it can yield negative squared rotor speeds, which are not physically meaningful. To ensure feasibility, the desired rotor speeds $\boldsymbol{\Omega}_{\text{des}}$ are obtained via a non-negative least squares (NNLS) problem
\begin{align}
\boldsymbol{x} = \arg\min_{\boldsymbol{x} \geq 0} \left\| \boldsymbol{u}_{\text{LQR}} - \mathbf{M} \, \boldsymbol{x} \right\|_2^2
\end{align}
followed by $\boldsymbol{\Omega}_{\text{des}} = \sqrt{\boldsymbol{x}}$. To ensure proper actuation, the $i$-th rotor speed is constrained within a lower limit \( \underline{\Omega} \), set by phase resistance and starting torque, and an upper limit \( \overline{\Omega} \), beyond which additional input no longer increases thrust, as given by 
\begin{align} \label{eq:constraints}
\Omega_{\text{in}, i} = \min\left( \, \overline{\Omega}, \, \max \left( \Omega_{\text{des}, i} , \underline{\Omega} \right) \right) \, .
\end{align}
Next, the microcontroller unit (MCU) translates the rotational speed commands into electrical signals by generating pulse-width modulation (PWM) signals to drive the motors. The low-level motor control pathway is modeled compactly as
\begin{align}
\boldsymbol{\Omega}_{\text{in}} = \mathcal{G}_1 ( \boldsymbol{u}_{\text{LQR}} ) \, .  \label{eq:tf_1}
\end{align}
%
\subsubsection{Power System} \label{sub:power}
Battery-powered aerial platforms, particularly those using Lithium-Polymer (LiPo) cells, are subject to high-current transients due to aggressive propulsion demands. These bursts give rise to nonlinear phenomena such as voltage sag, capacitive loading, and relaxation effects. To capture these time-varying behavior, the Thevenin-based equivalent circuit model is widely adopted \cite{bauersfeld2022range, zhang2020accurate, he2011evaluation}.
\\
Given that \( \Omega_{\text{in},i} \) is bounded by \eqref{eq:constraints}, and assuming a non-ideal rotor efficiency \( \eta_{\text{rot}} < 1 \), the electrical power drawn by the $i$-th motor scales cubically with speed as
\begin{align} \label{eq:power}
P_{\text{in},i} = \frac{\tau_{M,i} \, \Omega_{\text{in},i}}{\eta_{\text{rot}}} = \frac{k_M \, \Omega_{\text{in},i}^3}{ \eta_{\text{rot}} } \, .
\end{align}
Accordingly, the total current drawn from the battery to supply the combined motor power \( P_{\text{out}}(t) \) is then
\begin{align}
I(t) = { \sum_{i=1}^4 P_{\text{in},i}(t) } \, / \, {V(t)} \, , \label{eq:I_t}
\end{align}
%
In this framework, the terminal voltage \( V(t) \) is expressed as
\begin{align} 
V(t) = V_{\text{oc}}(t) - R_0 I(t) - V_1(t) \, , \label{eq:V_t}
\end{align}
where \( R_0 \) is the internal ohmic resistance, and $V_{\text{oc}}(t)$ is the open-circuit voltage (OCV), given by the quadratic fit
\begin{align}
V_{\text{oc}}(t) = \nu_0 + \nu_1 \, \text{SoC}(t) + \nu_2 \, \text{SoC}(t)^2 \, ,
\end{align}
with $\nu_i$'s being battery-specific constants listed in Appendix~\ref{appendix:b}.
The normalized state-of-charge (SoC $\in [0, 1]$), representing the remaining usable energy, evolves according to
\begin{align}
\dot{\text{SoC}}(t) = -\frac{I(t)}{C_{\text{bat}}} \, , \label{eq:SoC}
\end{align}
where \( C_{\text{bat}} \) is the nominal capacity. The voltage drop due to first-order electrochemical polarization dynamics follows
\begin{align}
\dot{V}_1(t) = -\frac{1}{\underbrace{R_1 C_1}_{\tau_{RC}}} V_1(t) + \frac{1}{C_1} I(t) \, , \label{eq:v_1}
\end{align}
where \( R_1 \) and \( C_1 \) characterizes the transient response time.
\\
Since the five variables involved in \eqref{eq:I_t}–\eqref{eq:v_1} are implicitly interdependent, the system constitutes a differential-algebraic equation (DAE), which is solved numerically at each time step to propagate the battery's internal dynamics as
\begin{align}
\boldsymbol{\Omega}_{\text{cmd}} = \mathcal{G}_2 ( \boldsymbol{\Omega}_{\text{in}} ) \, .  \label{eq:tf_2}
\end{align}
\subsubsection{Actuator Dynamics} 
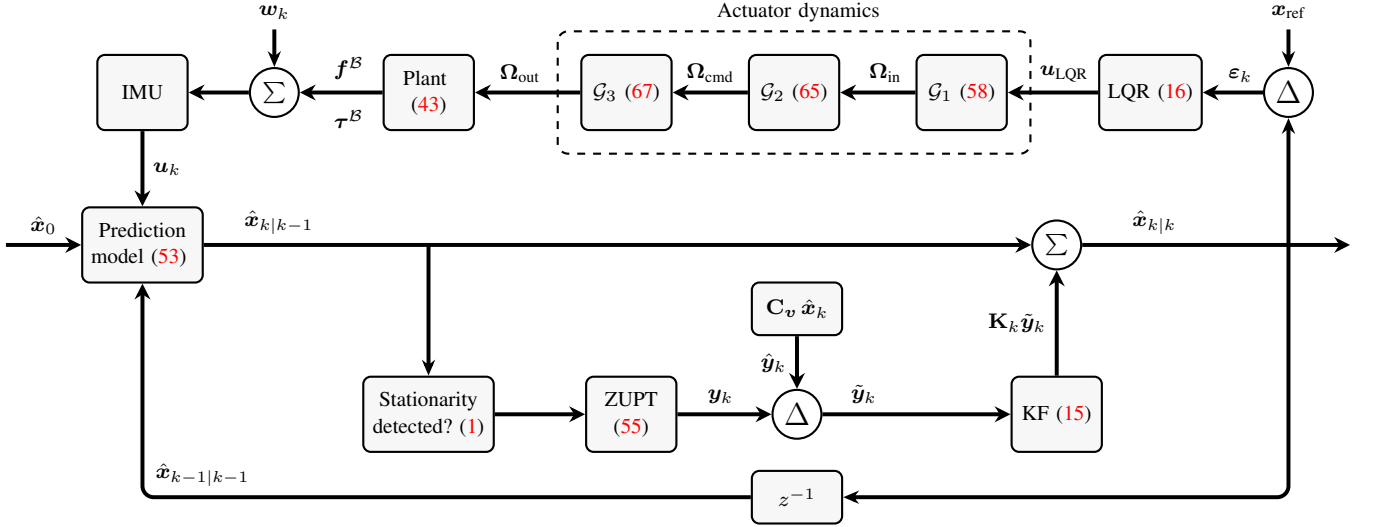
\begin{figure*}[t]
\centering
\begin{tikzpicture}[ 
  block/.style={draw, thick, minimum width=12mm, minimum height=1cm, fill=gray!6, rounded corners=1mm},
  block_small/.style={draw, thick, minimum width=12mm, minimum height=7mm, fill=gray!6, rounded corners=1mm},
  joint/.style={draw, circle, thick, minimum size=6.5mm, inner sep=0pt},
  arrow/.style={-{Stealth[length=6pt, width=6pt]}, rounded corners=1.5mm, line width=1.5pt},
  labelnode/.style={font=\footnotesize, inner sep=1pt},
  every node/.style={font=\footnotesize}
]

\node[block] (pred) {\shortstack{ \ Prediction\vspace{.5mm}\\model \eqref{eq:prediction}} \ };
\node[joint, right=110mm of pred] (J_sum_1) {$\sum$};
\node[block, below=14mm of J_sum_1] (kf) {KF \eqref{eq:LQE}};
\node[joint, left=25mm of kf] (J_sub_1) {\large $\Delta$};
\node[block_small, above=7mm of J_sub_1] (meas) {$\mathbf{C}_{\boldsymbol{v}} \, \hat{\boldsymbol{x}}_k$}; %
\node[block_small, below=4mm of J_sub_1] (pause) {$z^{-1}$};
\node[block, left=12.5mm of J_sub_1] (zupt) {\shortstack{ZUPT\vspace{.5mm}\\ \eqref{eq:v_zupt}}};
\node[block, align=center, left=12mm of zupt, text width=15mm] (stat) {\shortstack{Stationarity\\detected? \eqref{alg:zupt} }};
\node[block, above=10mm of pred] (imu) {IMU};
\node[joint, right=8mm of imu] (J_sum_2) {$\sum$};
\node[block, right=11mm of J_sum_2] (plant){\shortstack{Plant \vspace{.5mm}\\ \eqref{eq:x_dot}}}; 
\node[block, right=14mm of plant] (TF-3) {$\mathcal{G}_3$ \eqref{eq:tf_3}};
\node[block, right=10mm of TF-3] (TF-2) {$\mathcal{G}_2$ \eqref{eq:tf_2}};
\node[block, right=10mm of TF-2] (TF-1) {$\mathcal{G}_1$ \eqref{eq:tf_1}};
\node[block, right=12mm of TF-1] (lqr) {LQR \eqref{eq:LQR}};
\node[joint, right=8mm of lqr] (J_sub_2) {\large $\Delta$};

\draw[arrow] ([xshift=-10mm]pred.west) -- node[xshift=0mm, above] {$\hat{\boldsymbol{x}}_0$} (pred);
\draw[arrow] (pred) -- node[xshift=-45mm, above] {$\hat{\boldsymbol{x}}_{k|k-1}$} (J_sum_1);
\draw[arrow] (stat) -- node[xshift=-1.5mm, above] {} (zupt);
\draw[arrow] (zupt) -- node[xshift=-.5mm, above] {$\boldsymbol{y}_k$} (J_sub_1);
\draw[arrow] (meas) -- node[xshift=-0mm, left] {$\hat{\boldsymbol{y}}_k$} (J_sub_1);
\draw[arrow] (J_sub_1) -- node[xshift=-7mm, above] {$\tilde{\boldsymbol{y}}_k$} (kf);
\draw[<->, >={Stealth[length=6pt, width=5pt]}, line width=1.5pt, rounded corners=1.5mm](J_sub_2) |- (pause);
\draw[arrow] (pause) -| node[xshift=8mm, above] {$\hat{\boldsymbol{x}}_{k-1|k-1}$} (pred);
\draw[arrow] (kf) -- node[xshift=0mm, left] {$\mathbf{K}_k \tilde{\boldsymbol{y}}_k$} (J_sum_1);
\draw[arrow] (J_sum_1) -- ++(3.9,0) node[xshift=-26mm, above] {$\hat{\boldsymbol{x}}_{k|k}$};
\draw[arrow] (imu) -- node[right] {$\boldsymbol{u}_{k}$} (pred);
\draw[arrow] (J_sum_2) -- (imu);
\draw[arrow] (plant) -- node[xshift=1mm] {\shortstack{$\boldsymbol{f}^\mathcal{B}$\vspace{3mm}\\$\boldsymbol{\tau}^\mathcal{B}$}} (J_sum_2);

\draw[arrow] (TF-3) -- node[xshift=-1mm, above] {$\boldsymbol{\Omega}_{\text{out}}$} (plant);
\draw[arrow] (TF-2) -- node[xshift=0mm, above] {$\boldsymbol{\Omega}_{\text{cmd}}$} (TF-3);
\draw[arrow] (TF-1) -- node[xshift=1mm, above] {$\boldsymbol{\Omega}_{\text{in}}$} (TF-2);
\draw[arrow] (lqr) -- node[xshift=1.5mm, above] {$\boldsymbol{u}_{\text{LQR}}$} (TF-1);
\draw[arrow] (J_sub_2) -- node[xshift=1mm, above] {$\boldsymbol{\varepsilon}_{k}$} (lqr);

\draw[<-, >={Stealth[length=6pt, width=5pt]}, line width=1.5pt] (J_sub_2.north) -- ++(0,0.5) node[above] {$\boldsymbol{x}_{\text{ref}}$};
\draw[<-, >={Stealth[length=6pt, width=5pt]}, line width=1.5pt] (J_sum_2.north) -- ++(0,0.5) node[above] {$\boldsymbol{w}_{k}$};
\draw[<-, >={Stealth[length=6pt, width=5pt]}, line width=1.5pt] (stat.north) -- ++(0,1.75) node[above] {};

\draw[dashed, thick, rounded corners]($(TF-3.north west)+(-0.3,0.3)$) node[xshift=3.2cm, above]{Actuator dynamics} rectangle ($(TF-1.south east)+(0.3,-0.3)$);
\end{tikzpicture}
\caption{Signal flow schematic: near-stationarity detection (Alg.~\ref{alg:zupt}) triggers the C-ZUPT update via the inner bypass in \eqref{eq:v_zupt}.}
\label{fig:system}
\end{figure*}

The final stage models the airframe's dynamic response and its influence on the generated thrust and torques. While the control signal \( \boldsymbol{\Omega}_{\text{cmd}} \) represents the idealized instantaneous command, rotor inertia and aerodynamic effects introduce a time lag in the actual output speeds \( \boldsymbol{\Omega}_{\text{out}} \). 
\\
To capture this behavior, the rotor dynamics are modeled as a linear first-order system \cite{cook2012flight}
\begin{align}
\dot{\boldsymbol{\Omega}}_{\text{out}} = \frac{1}{\tau_{\text{rot}}} ( {\boldsymbol{\Omega}}_{\text{cmd}} - {\boldsymbol{\Omega}}_{\text{out}} ) \, ,
\end{align}
where \( \tau_{\text{rot}} \) denotes the dominant time constant.  
The output rotor speeds, which ultimately determine the generated forces and torques, are then governed by the dynamic mapping
\begin{align}
\boldsymbol{\Omega}_{\text{out}} = \mathcal{G}_3 ( \boldsymbol{\Omega}_{\text{cmd}} ) \, . \label{eq:tf_3}
\end{align}
To conclude, this chapter analyzes the linear hover characteristics, examines the real-time contribution of C-ZUPT—along with its heuristic basis and application—and ends with a model of the effective actuation output, encapsulated as
\begin{align}
\boldsymbol{\Omega}_{\text{out}} = \mathcal{G}_3 \big( \mathcal{G}_2\big( \mathcal{G}_1( \boldsymbol{u}_{\text{LQR}} )\big)\big) \, .
\end{align}
Fig.~\ref{fig:system} presents the discrete-time LQG signal iteration in this setup, emphasizing how the resulting rotor commands
\begin{align}
\boldsymbol{u}_{\text{out}} = \mathbf{M} \, \boldsymbol{\Omega}_{\text{out}}^{\circ 2} \, , \label{eq:u_out}
\end{align}
ultimately drive the system dynamics according to
\begin{align}
\dot{\boldsymbol{x}}(t) = \boldsymbol{f} \big(\boldsymbol{x}(t), \boldsymbol{u}_{\text{out}}(t) \big) \, .
\end{align}
\vspace{-2mm} 
%
%
\section{Results and Analysis} \label{sec:results}
This section presents the key technical metrics used to evaluate system performance. To address potential model inaccuracies in the KF framework, the left half of Table~\ref{t:cov_lqr_combined} lists the assumed process ($\mathbf{W}$) and measurement noise ($\mathbf{V}$) covariance matrices. For simplicity, these matrices are modeled as 3D isotropic ($\sigma_{ii} \boldsymbol{I}_3$), with numerical values detailed in Appendix~\ref{appendix:b}.
\\
Given the fast, underactuated nature of quadrotors, effective stabilization requires control corrections within milliseconds. As a result, the system operates near its stability margin, where even minor LQR mis-tuning can compromise performance.
\begin{table*}[b]
\centering
\caption{System design matrices: KF process and measurement covariances (left), LQR weights (right).}
\renewcommand{\arraystretch}{1.375} 
\begin{tabular}{|c|l|c|c|c||c|l|c|c|c|c|} \hline
\CC Symbol & Noise variable\CC & 3D Variance\CC & Units\CC & Dim.\CC &
Weight\CC & Variable\CC & Tolerance\CC & Result\CC & Units\CC & Dim.\CC \\
\hline
$\mathbf{W}_{\boldsymbol{\xi}}$ & Position & $\frac{1}{3}\sigma_f^2 \Delta t^3  $ & m$^2$ & \multirow{4}{*}{$\mathbb{R}^{3\times3}$} &
$\mathbf{Q}_{\boldsymbol{\xi}}$& Position & $\pm$0.1 & 100 & m$^{-2}$ & \multirow{4}{*}{$\mathbb{R}^{3\times3}$} \\ \cline{1-4} \cline{6-10}
$\mathbf{W}_{\boldsymbol{v}}$ & Velocity & $\sigma_f^2 \Delta t$ & (m/s)$^2$ & &
$\mathbf{Q}_{\boldsymbol{v}}$& Velocity & $\pm$0.2 & 25 & (m/s)$^{-2}$ & \\ \cline{1-4} \cline{6-10}
$\mathbf{W}_{\boldsymbol{\eta}}$ & Orientation & $\sigma_\omega^2 \Delta t$ & rad$^2$ & &
$\mathbf{Q}_{\boldsymbol{\eta}}$& Orientation & $\pm$0.1 & 100 & rad$^{-2}$ & \\ \cline{1-4} \cline{6-10}
$\mathbf{W}_{\boldsymbol{\omega}}$ & Body rates & $\sigma_{\omega}^2 / \Delta t$ & (rad/s)$^2$ & &
$\mathbf{Q}_{\boldsymbol{\omega}}$ & Body rates & $\pm$1.0 & 1 & (rad/s)$^{-2}$ & \\ 
\hline \hline
$\mathbf{V}_{\boldsymbol{\xi}}$ & GNSS Positioning & $\sigma_{_{\textit{GNSS}}}^2$ & m$^2$ & \multirow{4}{*}{$\mathbb{R}^{3\times3}$} &
$\mathrm{R}_{\boldsymbol{u}_z}$ & Thrusting & $m\text{g}/2$ & 4.752 & N$^{-2}$ & \multirow{4}{*}{$\mathbb{R}$} \\ \cline{1-4} \cline{6-10}
$\mathbf{V}_{\boldsymbol{v}}$ & Zero-update & $\sigma_{_{\textit{ZUPT}}}^2$ & (m/s)$^2$ & &
$\mathrm{R}_{\boldsymbol{u}_{\phi}}$ & Rolling & $\pm J_{xx} \dot{p}$ & 11.11 & (N$\cdot$m)$^{-2}$ & \\ \cline{1-4} \cline{6-10}
$\mathbf{V}_{\boldsymbol{\eta}}$ & Attitude sensing & $\sigma_{\eta}^2$ & rad$^2$ & &
$\mathrm{R}_{\boldsymbol{u}_{\theta}}$ & Pitching & $\pm J_{yy} \dot{q}$ & 11.11 & (N$\cdot$m)$^{-2}$ & \\ \cline{1-4} \cline{6-10}
$\mathbf{V}_{\boldsymbol{\omega}}$ & Rate sensing & $\sigma_{\textit{gyro}}^2$ & (rad/s)$^2$ & &
$\mathrm{R}_{\boldsymbol{u}_{\psi}}$ & Yawing & $\pm J_{zz} \dot{r}$ & 100 & (N$\cdot$m)$^{-2}$ & \\ \hline
\end{tabular} \label{t:cov_lqr_combined}
\end{table*}

\subsection{Performance metrics}
To address this sensitivity, the right side of Table~\ref{t:cov_lqr_combined} lists the state deviation ($\mathbf{Q}$) and control effort ($\mathbf{R}$) weighting matrices, selected using Bryson’s rule \cite{hespanha2018linear}, which specifies
\begin{align}
\mathbf{Q}_{ii} = \frac{1}{(x_{i,\text{max}})^2} \, , \quad \mathbf{R}_{jj} = \frac{1}{(u_{j,\text{max}})^2} \, .
\end{align}
This heuristic scales each weight based on the maximum expected deviation, while ensuring that cost function \eqref{eq:cost} is dimensionless by using inverse-squared physical units.
\\
The effectiveness of the controller is first assessed by measuring the time fraction it operates in saturation, defined as
\begin{align}
\tilde{u}_{\text{sat}} = \frac{1}{T} \int_0^T \mathbb{I}\left( | \boldsymbol{u}(t)| \geq \boldsymbol{u}_{\max} \right) \mathrm{d}t \, , \label{eq:u_sat}
\end{align}
where $\mathbb{I}(\cdot)$ denotes the indicator function. The energy expenditure is approximated by integrating the control effort, normalized by the nominal input vector \eqref{eq:equilibrium_input}, as follows
\begin{align}
\tilde{u}_{\text{tot}} = \frac{1}{T} \int_0^T \left\| {\boldsymbol{u}(t)} \oslash {\boldsymbol{u}_e} \right\|_2 \, \mathrm{d}t \, , \label{eq:u_tot}
\end{align}
where $\oslash$ denotes element-wise division. Finally, to quantify the evolution of system uncertainty, the trace of the error covariance matrix is normalized by its initial value, yielding
\begin{align}
\tilde{\zeta} (t) = \operatorname{tr}\left( \mathbf{P} (t) \right) / \operatorname{tr}(\mathbf{P}_{0}) \, . \label{eq:zeta}
\end{align}
\newpage

\subsection{Hovering Analysis} \label{sub:hover}
\begin{figure}[h] 
\begin{center}
\includegraphics[width=0.48\textwidth]{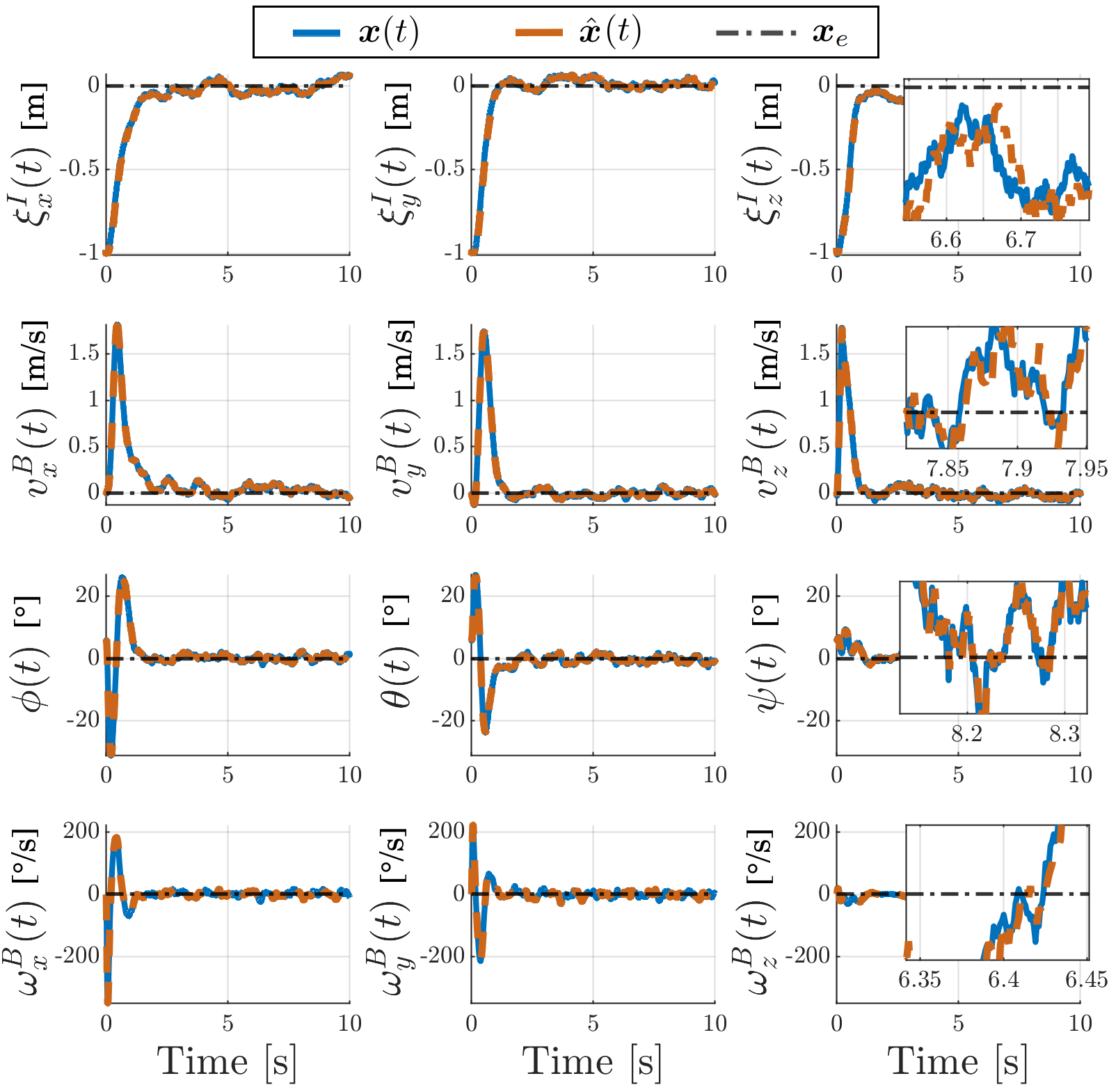}
\caption{Full-state trajectories with continuous updates ($\Gamma=1$).}
\label{fig:states_1}
\end{center}
\end{figure}

\begin{figure}[b] 
\begin{center}
\includegraphics[width=0.46\textwidth]{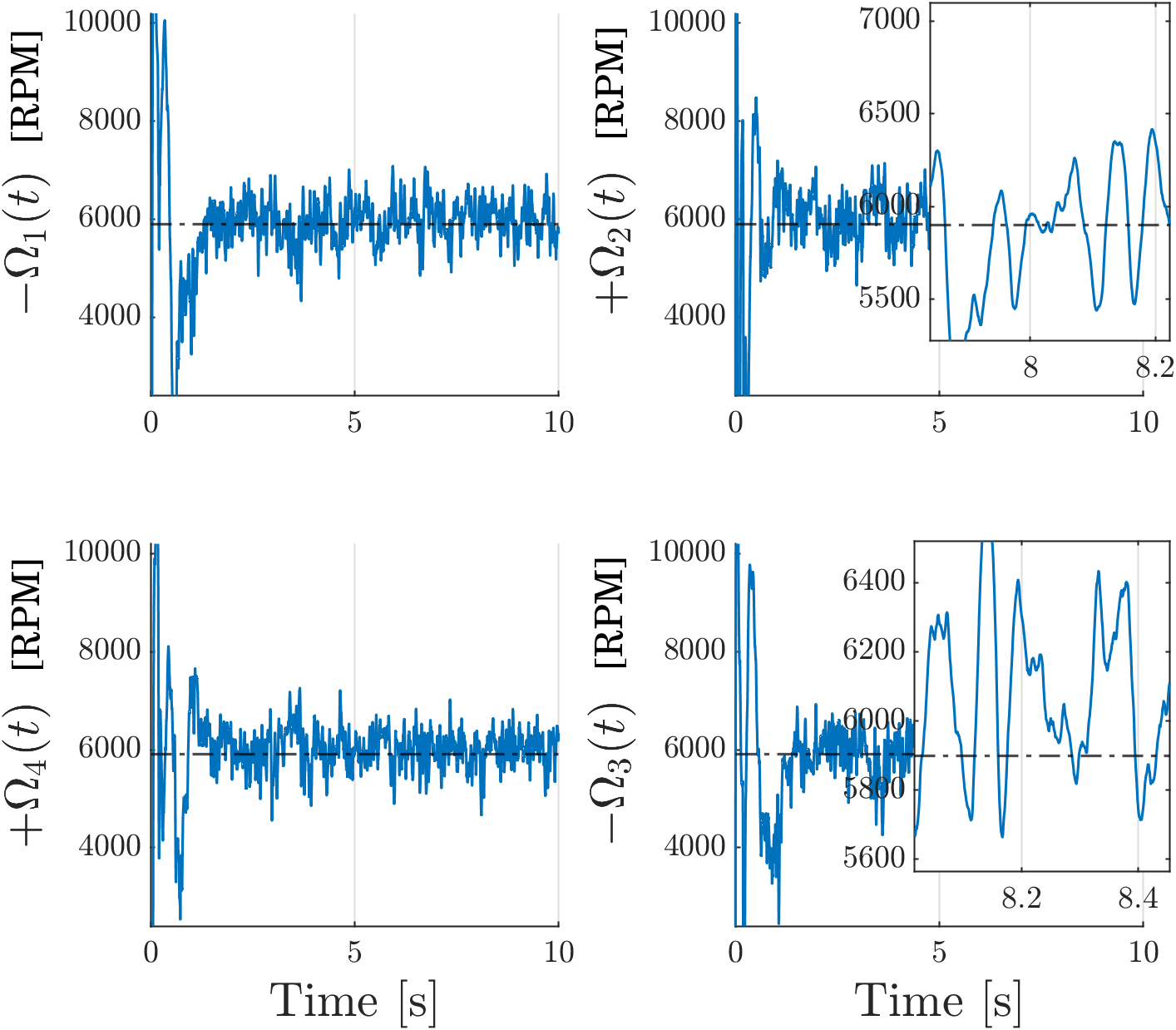}
\caption{Control mixer output: four motor speed commands.}
\label{fig:RPM_1}
\end{center}
\end{figure}

\begin{figure}[t] 
\begin{center}
\includegraphics[width=0.48\textwidth]{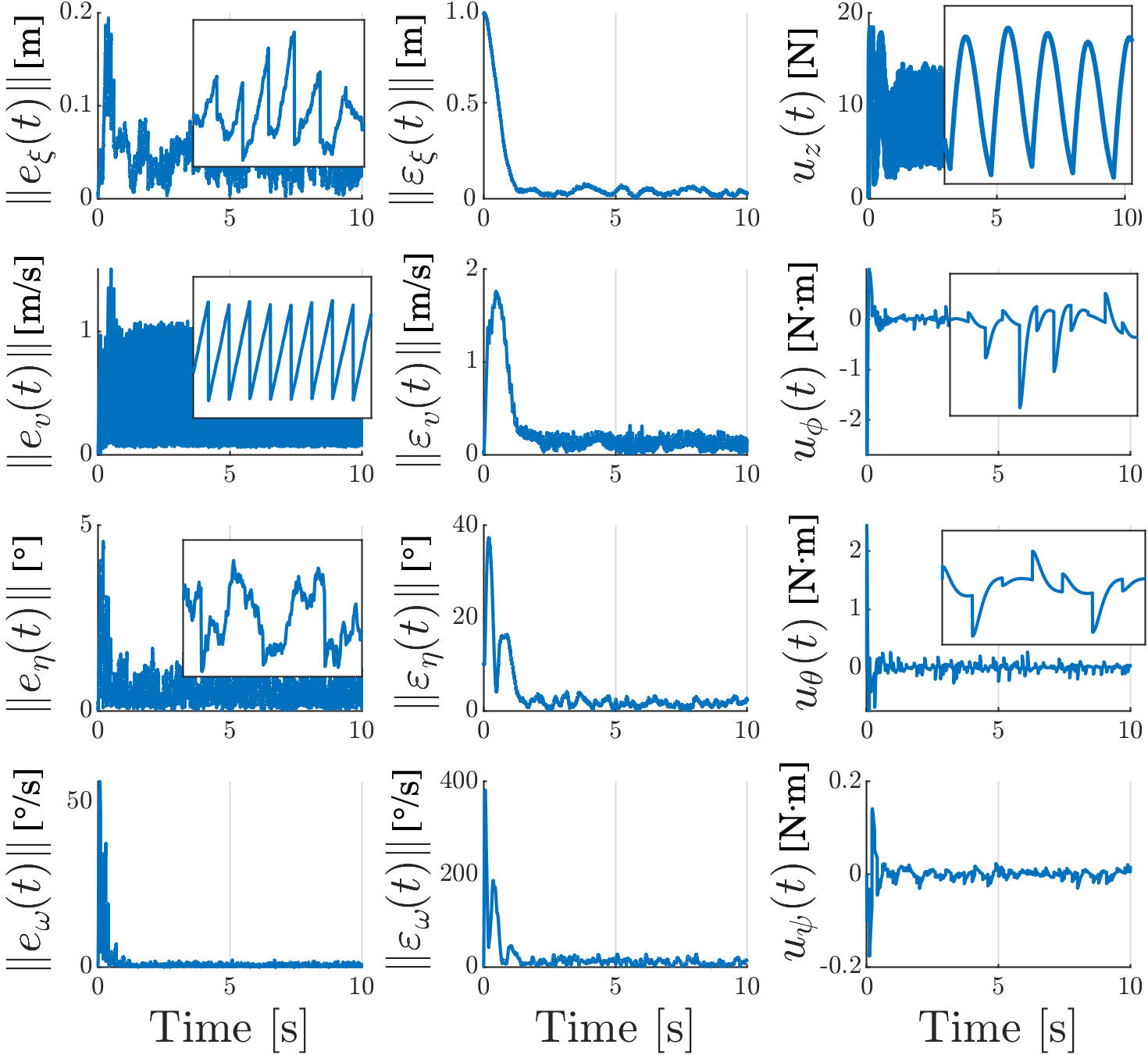}
\caption{Error norm evolution for infrequent updates ($\Gamma=0.05$).}
\label{fig:states_2}
\end{center}
\end{figure}

First, we analyze the stabilization capabilities of the LQG controller during steady hovering, using non-equilibrium initial conditions ($\boldsymbol{x}_0 \neq \boldsymbol{x}_e$) and process noise to emulate wind. For this baseline case, the position update frequency \eqref{eq:obs_1} is set equal to the prediction rate ($\Gamma = 1$). Fig.~\ref{fig:states_1} shows the resulting state trajectories over a ten-second interval.
\begin{figure}[b] 
\begin{center}
\includegraphics[width=0.465\textwidth]{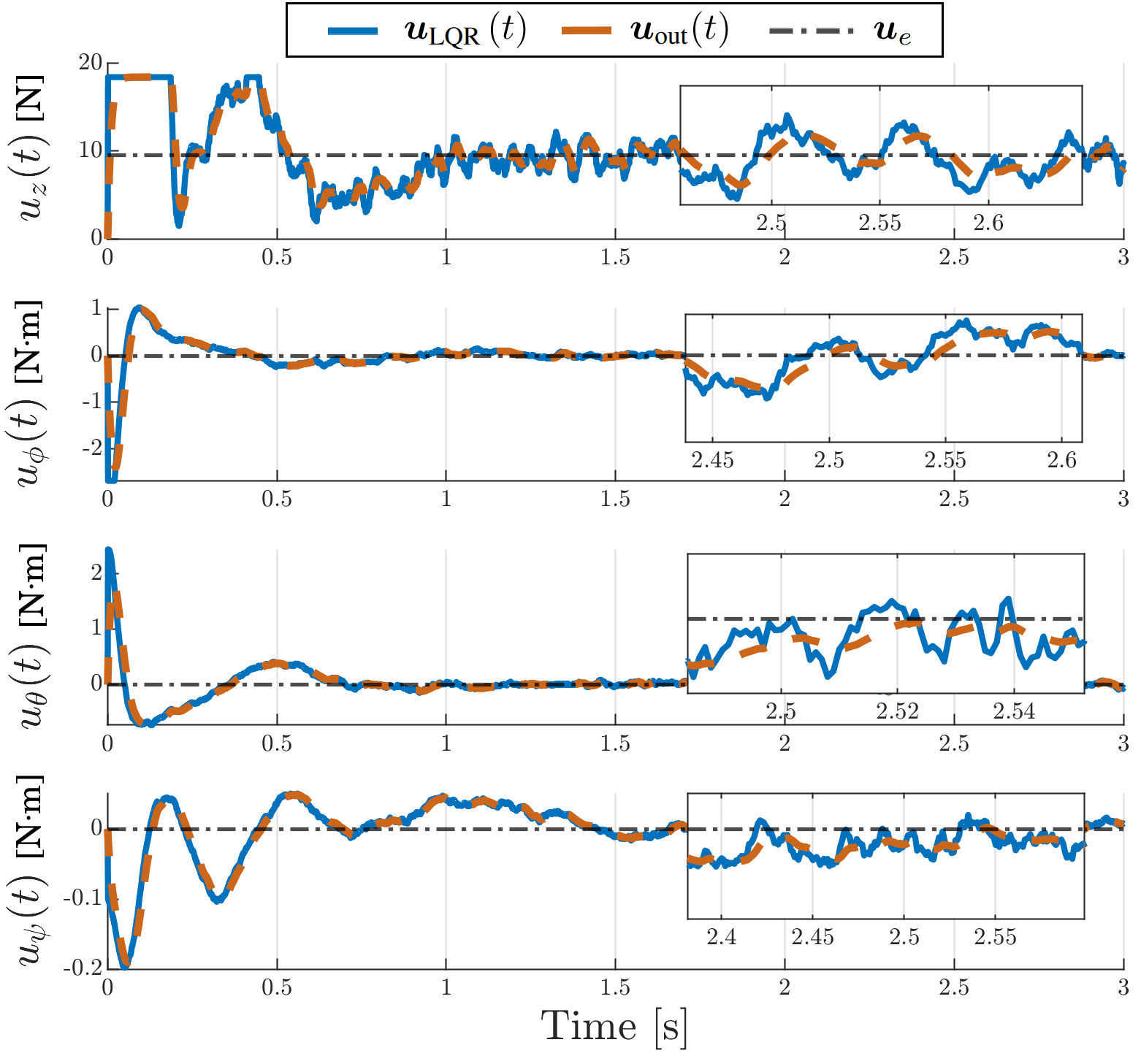}
\caption{Zoomed-in view of LQG-generated motor commands.}
\label{fig:control_1}
\end{center}
\end{figure}
\\
The subplots, arranged from top to bottom, show position coordinates, body-frame velocities, Euler angles, and body angular rates. For clarity, the state equilibrium point \eqref{eq:equilibrium_state} is set at the origin, $\boldsymbol{x}_e := \mathbf{0}_{12}$, and is marked by a dashed line. 
\\
As this scenario shows, despite adverse initial conditions and substantial process noise, the chosen $\mathbf{Q}$ and $\mathbf{R}$ weightings, combined with continuous updates ($\Gamma = 1$), keep the system well-regulated across all 12 state variables. The magnified insets further highlight the KF performance, showing that the state estimates remain tightly aligned with the true states.
\\
Fig.~\ref{fig:control_1} shows a close-up of the four control inputs used for stabilization. Solid lines show the LQR input commands, while dashed brown lines depict the actual outputs \eqref{eq:u_out}, demonstrating minimal phase lag between them. The desired setpoint corresponds to the hovering equilibrium \eqref{eq:equilibrium_input}, marked by a black dashed line, displaying steady-state reference values for thrust (top) followed by roll, pitch, and yaw torques.
\\
From a numerical standpoint, sustaining a hover means that the total thrust must match the force of gravity, so that vertical acceleration remains zero. Thus,
\begin{align}
\boldsymbol{T}^{\mathcal{B}} = \sum_{i=1}^4 \, f_{z,i} \, \textbf{e}_z^{\mathcal{B}} =  m \, \textbf{\textit{g}}^{\mathcal{B}} \, ,
\end{align}
where each rotor's vertical thrust is modeled as
\begin{align} \label{eq:hovering}
f_{z,i} = \frac{1}{2} \, \rho \underbrace{A_r}_{(\pi r^2)} C_T \underbrace{\bar{v}_z^2}_{ ( \Omega_i r )^2 } = k_T \, \Omega_{i}^2 \, ,
\end{align}
where $k_T$ is a lumped thrust coefficient that combines the effects of air density $\rho$, rotor disk area $A_r$, thrust coefficient $C_T$, and rotor tip speed $\Omega_i r$ \cite{cook2012flight}. 
%
By substituting numerical values into \eqref{eq:hovering}, the $i$-th baseline rotor speed required to maintain steady hover is approximated as
\begin{align}
\Omega_{\text{hov},i} \, \geq \, \sqrt{\frac{m\textit{g}}{4 \, k_T}} \approx  5{,}921 \ [\mathrm{RPM}] \, . \label{eq:RPM}
\end{align}
Fig.~\ref{fig:RPM_1} shows the actual rounds-per-minute (RPM) speeds commanded by the speed controller, computed by inverting the control outputs via the mixer matrix \eqref{eq:mixer_matrix}. For reference, black dashed lines mark the steady-state hover RPM from \eqref{eq:RPM}, with positive values indicating counter-clockwise rotation.
\\
As is common in GNSS-denied or indoor environments, limited measurement availability reduces the position update rate ($\Gamma < 1$), forcing the controller to rely more heavily on state predictions to maintain stability.
\\
Fig.~\ref{fig:states_2} illustrates this scenario for $\Gamma = 0.05$, implying that a position update occurs once every 20 prediction steps. To aid interpretation, the figure retains the same vertical arrangement as before and projects the 3D state vectors into L2 norms, thereby showing their error magnitudes. The left column depicts the estimation error \eqref{eq:err_est}, the center column shows the control error \eqref{eq:err_ctr}, and the right column presents the stabilization effort through all four control inputs.
\setcounter{figure}{9}
\begin{figure}[b] 
\begin{center}
\includegraphics[width=0.48\textwidth]{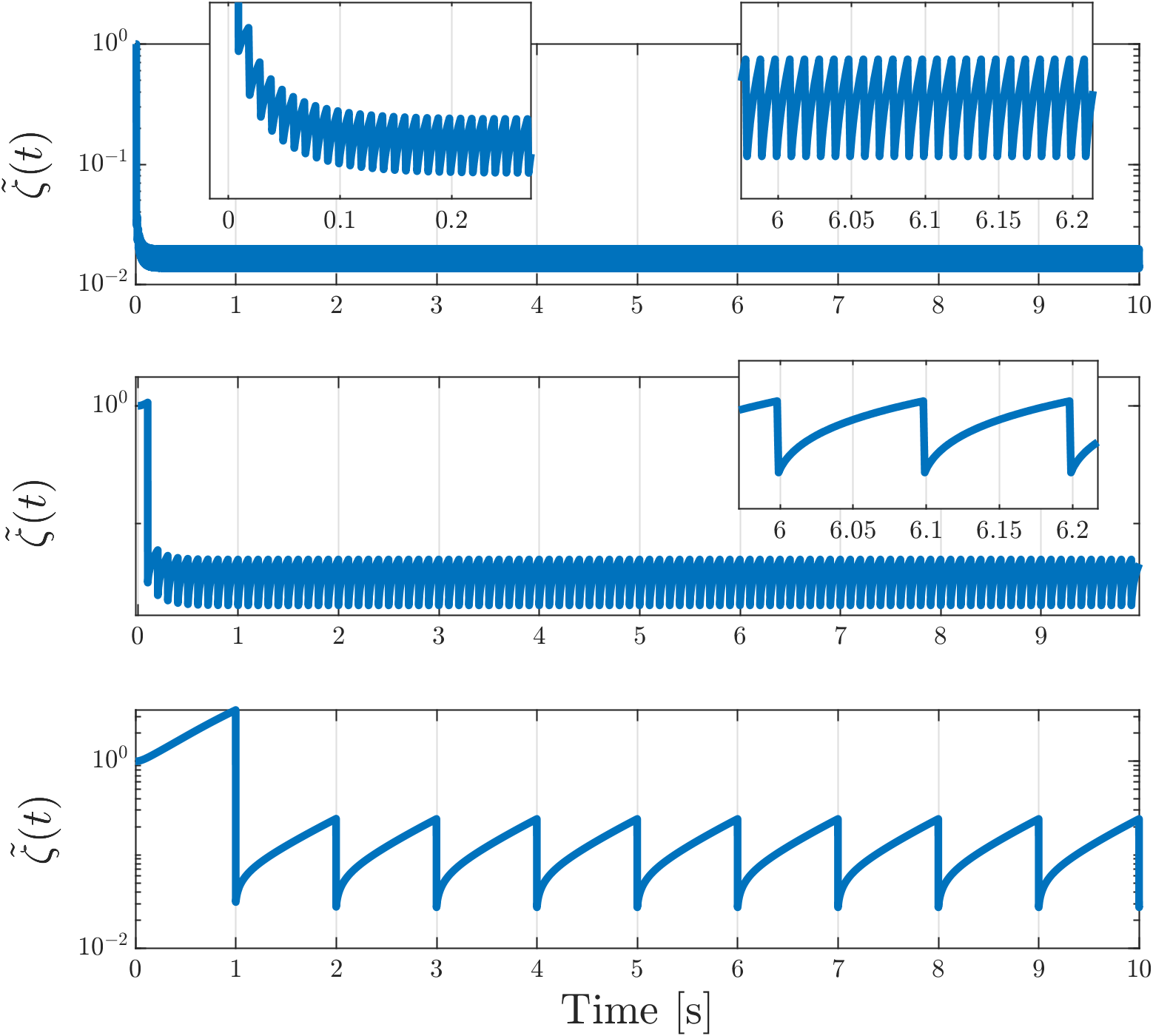}
\caption{Normalized total uncertainty for $\Gamma=\{1, 0.05, 0.001\}$.}
\label{fig:DoO_1}
\end{center}
\end{figure}
\\
As visible in the zoomed-in insets, the reduced update rate induces a random-walk effect in the estimation error, producing a sawtooth-like pattern whose local minima correspond to the periodic corrections. Despite this, the controller successfully keeps the system stable, albeit with increased noise.
\\
Fig.~\ref{fig:states_3} extends this analysis as the update rate drops to $\Gamma=0.001$, showing that both estimation and control errors, $\boldsymbol{e}$ and $\boldsymbol{\varepsilon}$, respectively, diverge markedly, while the controller oscillates erratically within its bounds.
\setcounter{figure}{8}
\begin{figure}[t]
\begin{center}
\includegraphics[width=0.4875\textwidth]{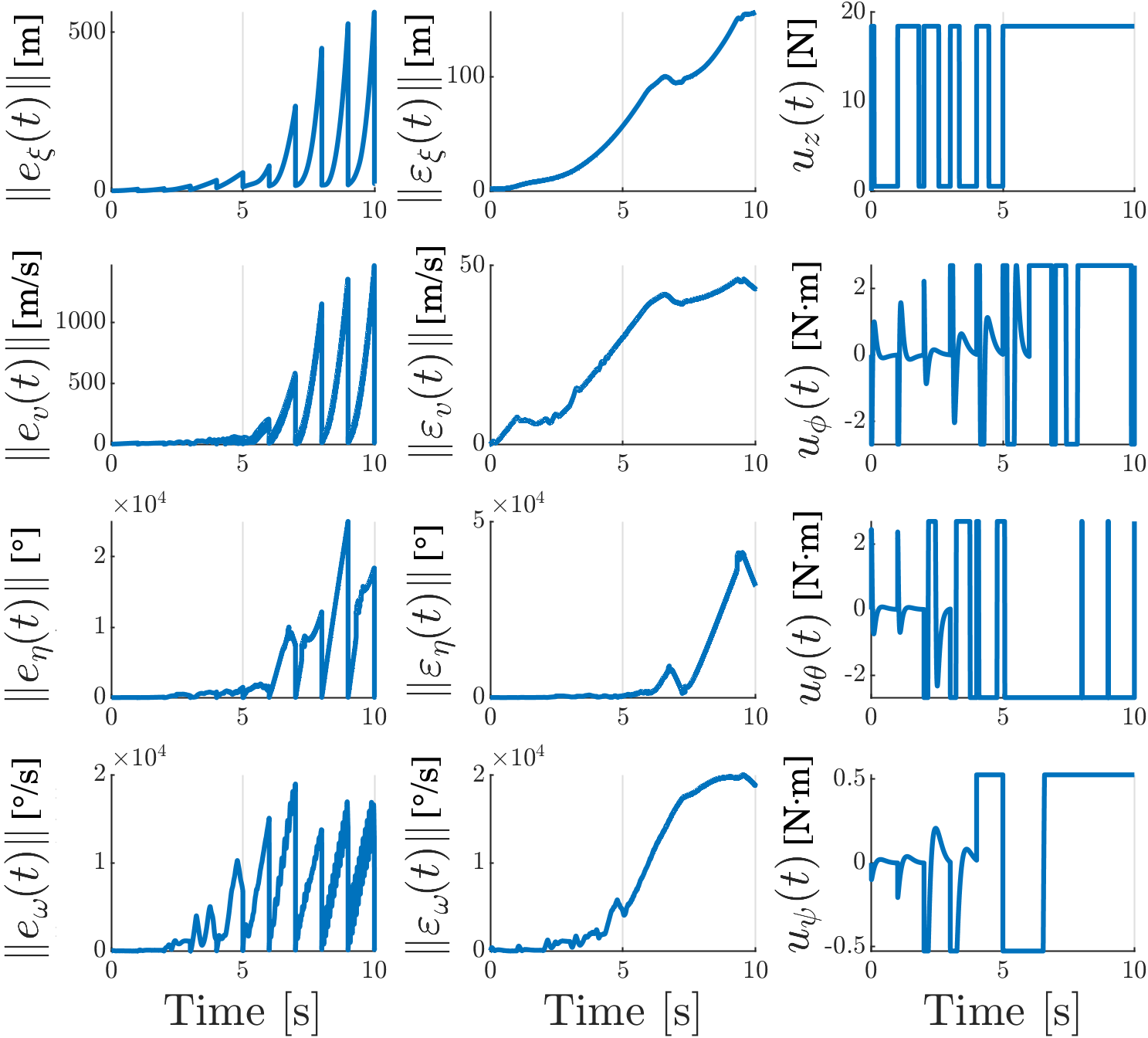}
\caption{Error norm evolution for sparse updates $\Gamma=0.001$.}
\label{fig:states_3}
\end{center}
\end{figure}
\\
Fig.~\ref{fig:DoO_1} offers an intuitive view of system uncertainty based on \eqref{eq:zeta}, with each curve normalized to start at 1.0. As process noise accumulates, the uncertainty growth rate depends on $\Gamma$, which controls how often corrections are introduced. Thus, a lower $\Gamma$ permits increased random walk effects, leading to wider error bands and degraded system observability.
\begin{table}[b]
\centering
\caption{Empirical assessment of LQG stability vs $\Gamma$.}
\renewcommand{\arraystretch}{1.65}
\begin{tabular}{|c|c|c||c|c||c|}
\hline 
$\Gamma$ \CC & $\| \boldsymbol{\varepsilon}_{\boldsymbol{ \xi}^{\mathcal{I}}} \|$ [m] \CC & $\| \boldsymbol{ \varepsilon}_{ \boldsymbol{\eta}^{\mathcal{I}}} \|$ [deg] \CC& $\tilde{u}_{\text{sat}}$ \CC  & $\tilde{u}_{\text{tot}}$ \CC & $\tilde{\zeta}_{ss}$  $(\mu \pm \sigma)$\CC \\ \hline 
1.0 & 0.021 & 0.745 & 0.0 & 1.011 & 0.02 $\pm$ 0.01 \\ \hline 
0.05 & 0.092 & 2.213 & 0.0 & 3.021 & 0.04 $\pm$ 0.02 \\ \hline 
0.01 & 0.311 & 3.177 & 0.08 & 12.314 & 0.06 $\pm$ 0.03 \\ \hline 
0.005 & 1.760 & 12.09 & 0.12 & 37.88 & 0.09 $\pm$ 0.06 \\ \hline 
0.001 & $\to \infty$ & $\to \infty$ & $\to$ 1 & $\to \infty$ & 0.14 $\pm$ 0.08 \\ \hline 
\end{tabular} \label{t:err_1}
\end{table} 
\\
Table~\ref{t:err_1} summarizes the key findings of this section, focusing on the norms of trajectory and attitude errors, $\boldsymbol{\varepsilon}_{\boldsymbol{\xi}^{\mathcal{I}}}$ and $\boldsymbol{\varepsilon}_{\boldsymbol{\eta}^{\mathcal{I}}}$, evaluated at the final time step $T$. 
\\
Based on the metrics in \eqref{eq:u_sat}–\eqref{eq:zeta}, the high demand for agile control pushes the LQG controller near its stability margin, revealing its sensitivity to estimator performance: while moderate reductions in $\Gamma$ can be offset by increased control effort, further reductions quickly erode stability as the estimates lag behind the fast dynamics, ultimately diverging.


\subsection{C-ZUPT Analysis}
As outlined in Algorithm~\ref{alg:zupt}, the stationarity check relies on specific force data and is therefore inherently limited by accelerometer noise. Thereby, motion sensitivity depends on proper tuning of two key parameters: the threshold levels ($\delta$), which set the allowable deviation, and the window size ($K$), which sets the averaging span.
\\
Fig.~\ref{fig:zupt_1} depicts a 7-second close-up of the detection mechanism in action, with specific forces on the left (blue), velocity estimates on the right (red), and corresponding thresholds indicated by dashed lines below. Initially, the quadrotor applies higher control gains to counteract larger control errors, then gradually reduces them as it settles near the reference point. 
\setcounter{figure}{10}
\begin{figure}[h] 
\begin{center}
\includegraphics[width=0.48\textwidth]{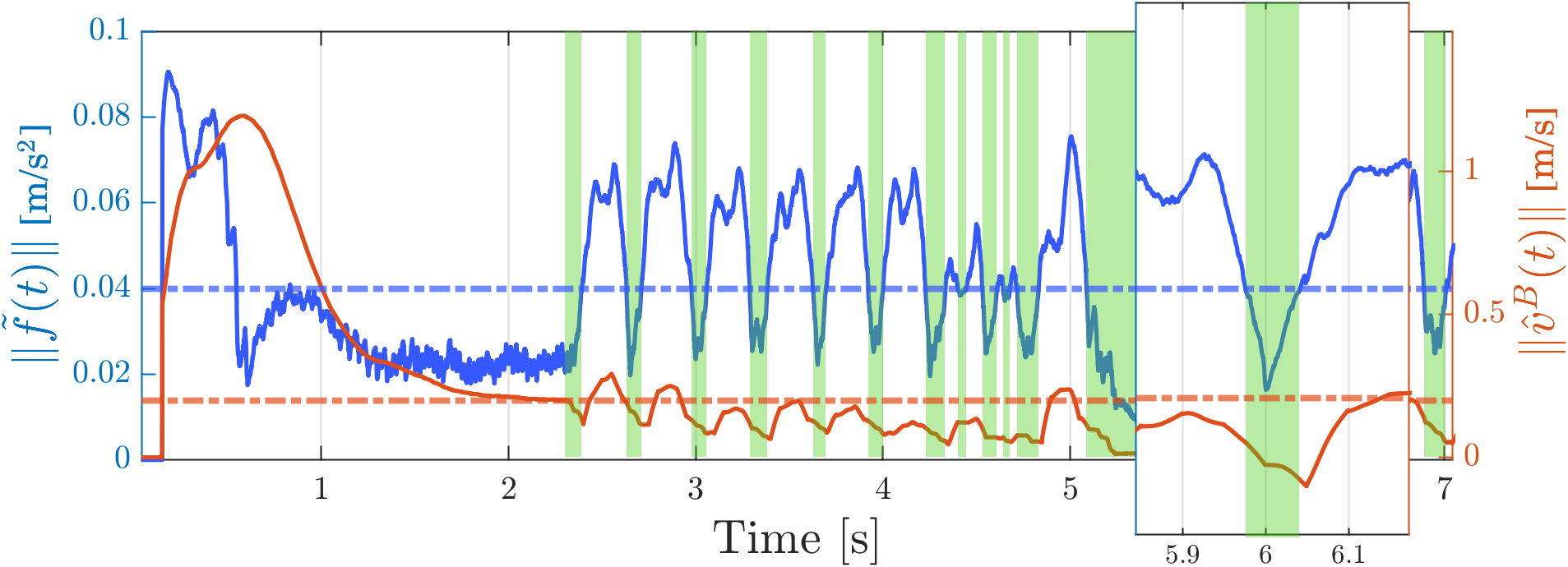}
\caption{Detection performance under strict thresholding.}
\label{fig:zupt_1}
\end{center}
\end{figure}
\\
The enlargement shows that detection (in pale green) is confirmed only when both measurable norms simultaneously fall below their respective thresholds. To further illustrate how heuristic settings affect sensitivity, Fig.~\ref{fig:zupt_2} presents a scenario with larger threshold ($\delta$) and window size ($K$) values. 
\setcounter{figure}{11}
\begin{figure}[h] 
\begin{center}
\includegraphics[width=0.48\textwidth]{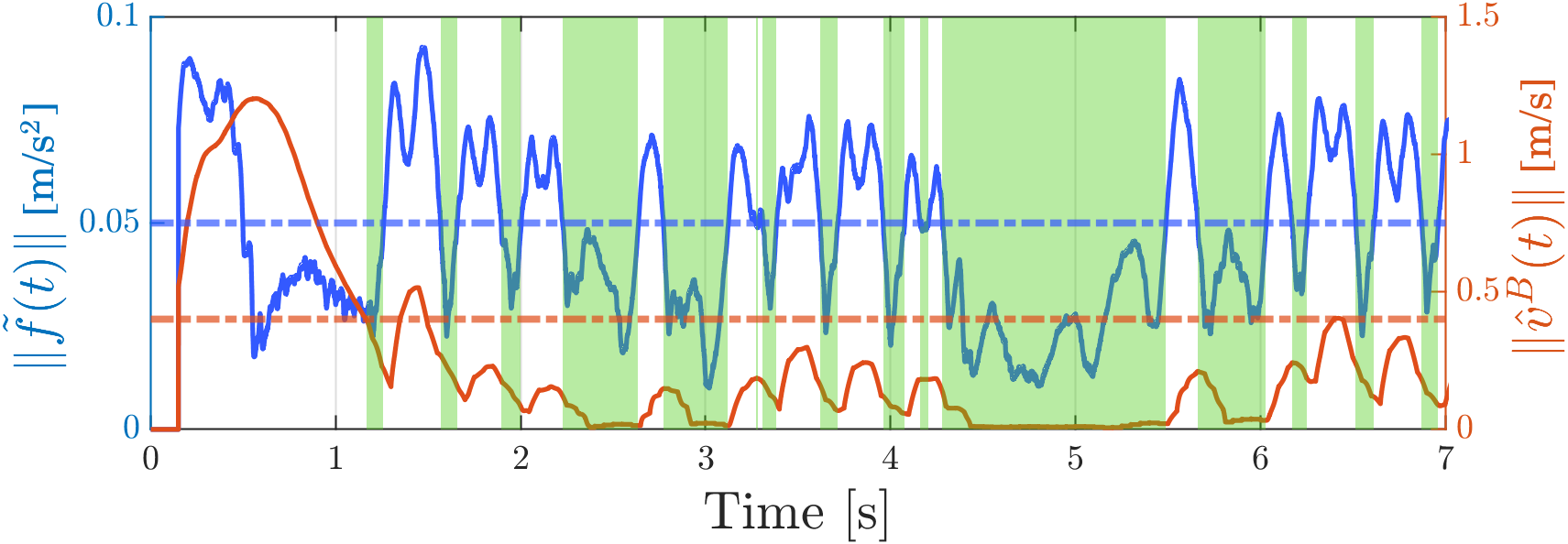}
\caption{Detection performance under permissive thresholding.}
\label{fig:zupt_2}
\end{center}
\end{figure}
\\
The dense green background indicates that classifying higher dynamics stationary increases the risk of false positives—especially during brief transients with high measurement uncertainty—underscoring the need for careful tuning.
\setcounter{figure}{12}
\begin{figure}[b]
\begin{center}
\includegraphics[width=0.48\textwidth]{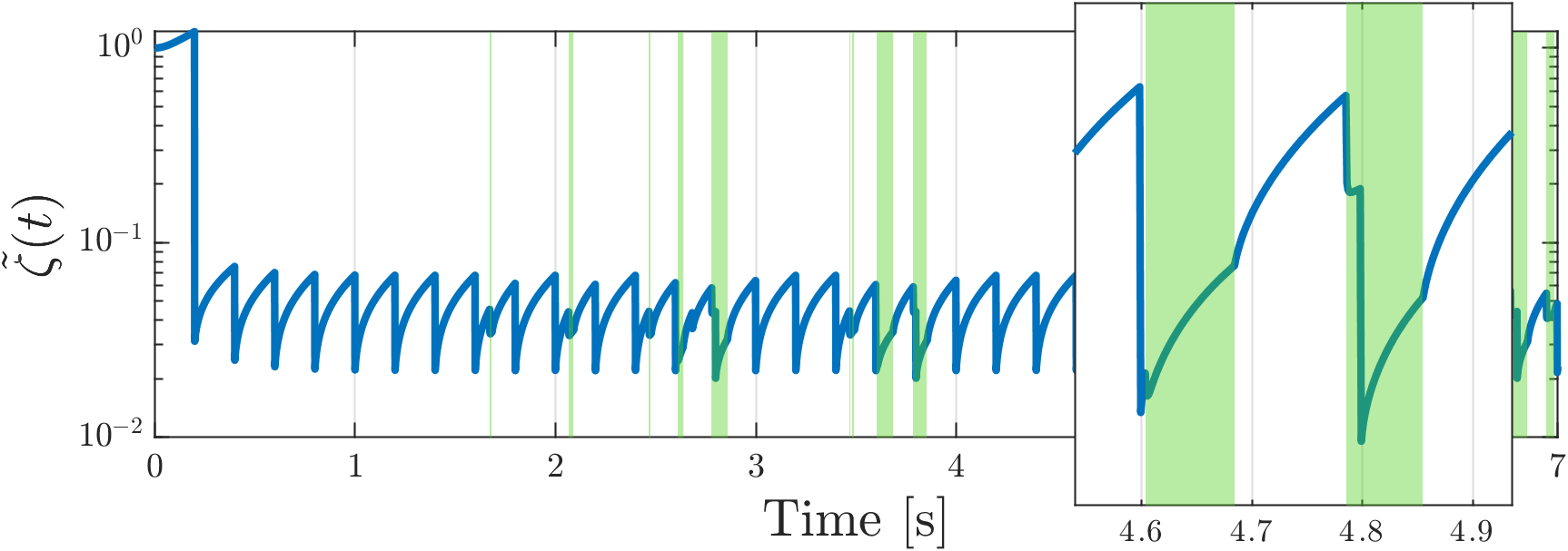}
\caption{Normalized total uncertainty with C-ZUPT ($\Gamma$=0.005).}
\label{fig:DoO_2}
\end{center}
\end{figure}
Lastly, Fig~\ref{fig:DoO_2} displays the real-time contribution of C-ZUPT to the normalized uncertainty metric \eqref{eq:zeta}.
\\
\setcounter{figure}{13}
\begin{figure}[t]
\begin{center}
\includegraphics[width=0.5\textwidth]{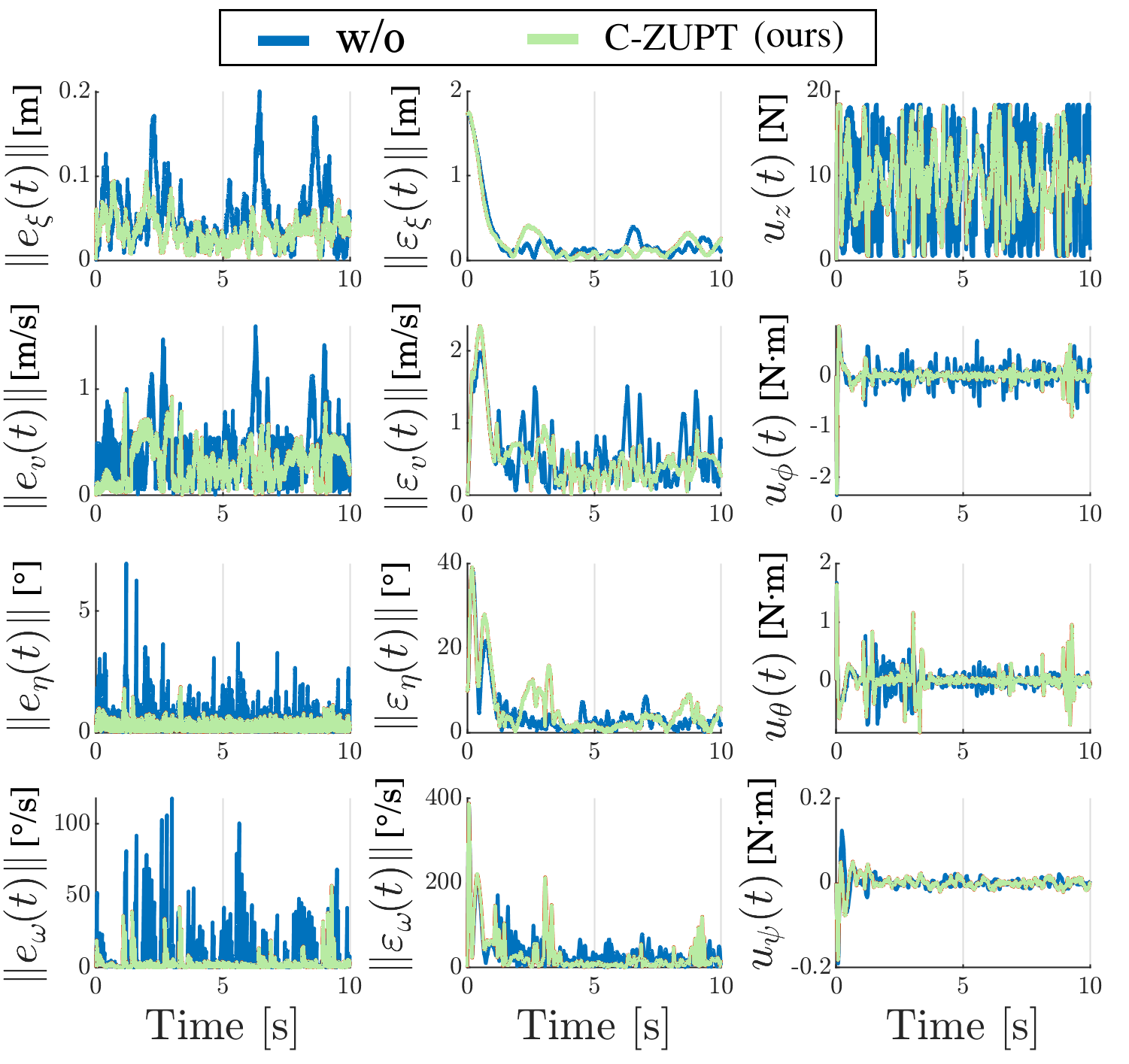}
\caption{Head-to-head comparison: unaided vs C-ZUPT-aided.}
\label{fig:zupt_err}
\end{center}
\end{figure}
In this example, position updates occur once every 200 steps, allowing estimation uncertainty to grow significantly between corrections, as reflected by the sawtooth peaks. However, when near-stationarity windows are detected, ZUPT corrections not only reset the accumulated error but also limit drift, giving a clear advantage over open-loop dead reckoning.
\\
To address the key question—Can the C-ZUPT-aided execution outperform the unaided one (w/o)?—Fig.~\ref{fig:zupt_err} shows a head-to-head comparison using the same error analysis as prior. 
With infrequent updates, estimability quickly deteriorates due to inertial sensor noise and the platform’s inherently erratic dynamics. This is evident in the unaided case (blue), where errors grow unchecked, demanding aggressive control actions. In contrast, the C-ZUPT cases (pale green) consistently maintain lower error magnitudes by exploiting detected stationarity periods that yield informative corrections.
\\
Table~\ref{t:err_2} summarizes the C-ZUPT performance, showing metrics normalized relative to unaided results over ten runs. Except for the bottom row, where both diverge, C-ZUPT consistently improves as $\Gamma$ decreases, yielding lower tracking error (28.8\%), attitude error (21.6\%), saturation time (76.9\%), control effort (31.9\%), and estimation uncertainty (28\%).
\begin{table}[b]
\centering
\caption{C-ZUPT metrics normalized to unaided results.}
\renewcommand{\arraystretch}{1.65}
\begin{tabular}{|c|c|c||c|c||c|}
\hline 
$\Gamma$ \CC & $\| \Bar{\boldsymbol{\varepsilon}}_{\boldsymbol{ \xi}^{\mathcal{I}}} \|$ \CC & $\| \Bar{\boldsymbol{ \varepsilon}}_{\boldsymbol{\eta}^{\mathcal{I}}} \|$ \CC & $\tilde{u}_{\text{sat}}$ \CC  & $\tilde{u}_{\text{tot}}$ \CC & $\tilde{\zeta}_{ss}$ $(\Bar{\mu} \, , \Bar{\sigma} )$ \CC \\ \hline 
0.1 & 0.982 & 0.979 & $-$ & $\approx$ 1.0  & (0.97, 0.98) \\ \hline 
0.05 & 0.951 & 0.923 & $-$ & 0.878 & (0.92, 0.95)\\ \hline 
0.01 & 0.834 & 0.840 & 0.427 & 0.774 & (0.87, 0.91) \\ \hline 
0.005 & 0.712 & 0.784 & 0.231 & 0.681 & (0.72, 0.88) \\ \hline 
0.001 & $-$ & $-$ & $\to$ 1 & $-$ & $-$ \\ \hline 
\end{tabular} \label{t:err_2}
\end{table} 
\newpage

\subsection{Power Analysis}
Having confirmed the positive effect of C-ZUPT on LQG performance, the next step is to examine its impact on instantaneous power consumption. To estimate the nominal power required for hovering, the steady-state RPM given in \eqref{eq:RPM} is substituted into the power model \eqref{eq:power}, yielding
\begin{align}
P_{\text{hov}} = \sum_i^4 P_{\text{in},i} = \frac{k_M}{2 \, \eta_{\text{rot}}} \left( \frac{m\text{g}}{k_T} \right)^{3/2} = 78.55 \ [\mathrm{W}] \, .
\end{align}
To evaluate how this energy demand constrains the operational envelope, the DAE system described in Sec.~\ref{sub:power}  solved assuming a standard 4S LiPo battery with a nominal voltage of 14.8 $\mathrm{V}$, (3.7 $\mathrm{V}$/cell), as detailed in Table~\ref{t:battery_1}.
\begin{table}[h]
\centering
\caption{Nominal hovering time based on battery specs.}
\renewcommand{\arraystretch}{1.4}
\begin{tabular}{|c|l|c|c|} \hline
\CC Symbol & Description\CC & Value\CC & Units\CC \\ \hline
$I_{\text{hov}}$ & Average current draw in hover & 5.31 & $\mathrm{A}$ \\ 
$V_{\text{hov}}$ & Terminal voltage in hover & 14.8 & $\mathrm{V}$ \\ 
$C_{\text{bat}}$ & Rated battery capacity & 3.0 & $\mathrm{Ah}$ \\ 
$E_{\text{bat}}$ & Approximate available energy & 44.4 & $\mathrm{Wh}$ \\ \hline \hline
$t_{\text{ideal}}$ & Ideal hover time (100\% capacity)  & $\approx$ 33.6 &  \multirow{2}{*}{mins} \\ 
$t_{\text{safe}}$ & \textbf{Recommended hover time} (70\% usable) & $\approx$ \textbf{23.5} & \\ \hline
\end{tabular} \label{t:battery_1}
\end{table}
\\
Under nominal conditions, the battery can sustain about 23.5 minutes of flight before hitting the 30\% safe margin. However, when wind disturbances or GNSS dropouts occur, the system demands more control to maintain stability, leading to higher power consumption and faster battery depletion.
\setcounter{figure}{14}
\begin{figure}[b] 
\begin{center}
\includegraphics[width=0.47\textwidth]{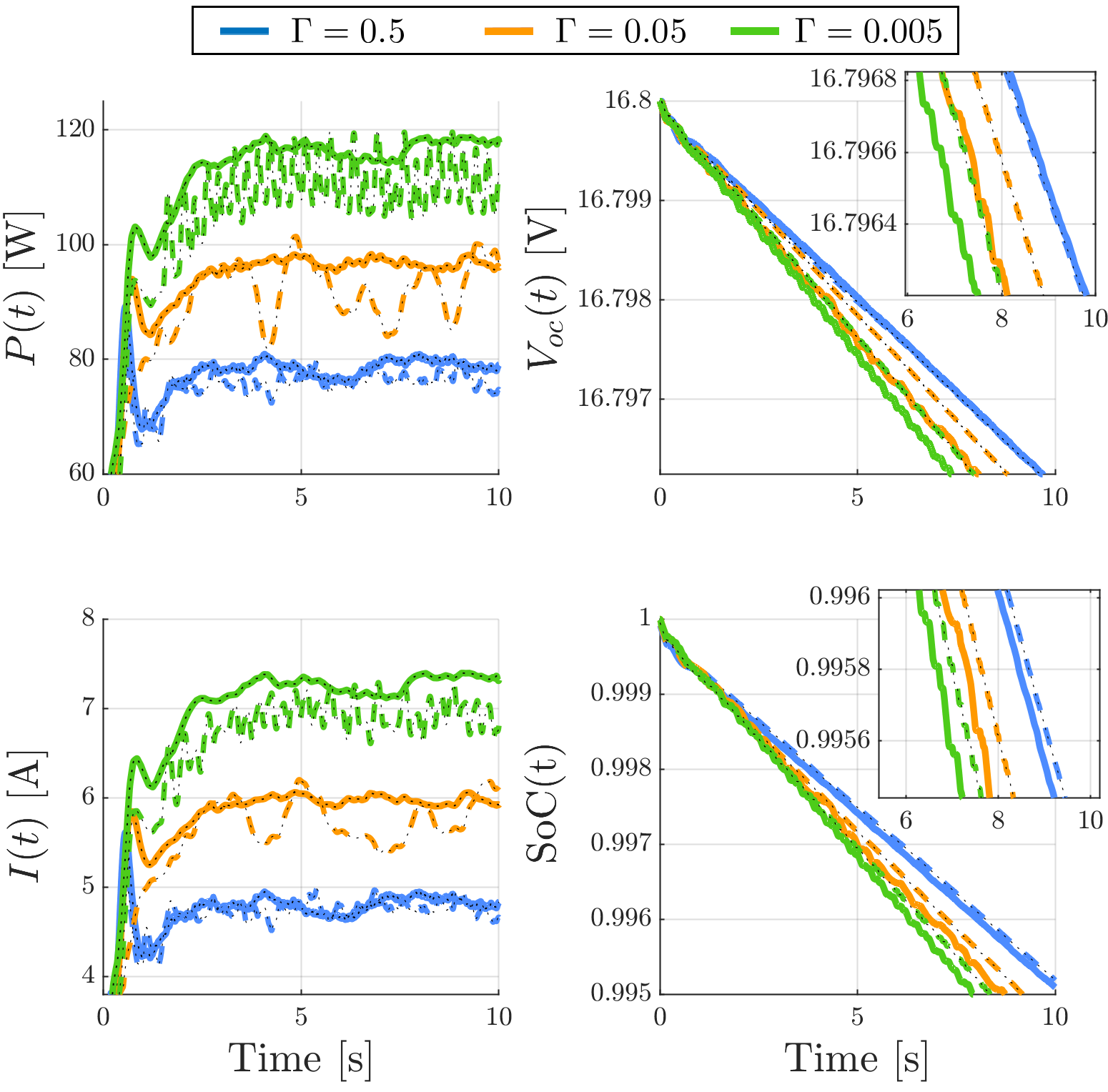}
\caption{Battery states: 10-second interval comparison.}
\label{fig:power_1}
\end{center}
\end{figure}
\\
Fig.~\ref{fig:power_1} illustrates the battery’s transient response to a unit disturbance over a 10-second interval, showing the time evolution of power, open-circuit voltage (OCV), state of charge (SoC), and current, arranged in clockwise order.
\\
These subplots highlight several key points. First, regardless of whether C-ZUPT is applied (dashed) or not (solid), all electrical states evolve as time-dependent responses to power consumption. Second, stronger disturbances (i.e., lower $\Gamma$) demand greater stabilization effort, evident in the left half of the figure, leading to a faster voltage drop and accelerated battery depletion on the right. Third, the C-ZUPT-aided cases exhibit slightly lower energy consumption than the unaided baseline, as highlighted in the magnified insets.
\begin{figure}[t] 
\begin{center}
\includegraphics[width=0.47\textwidth]{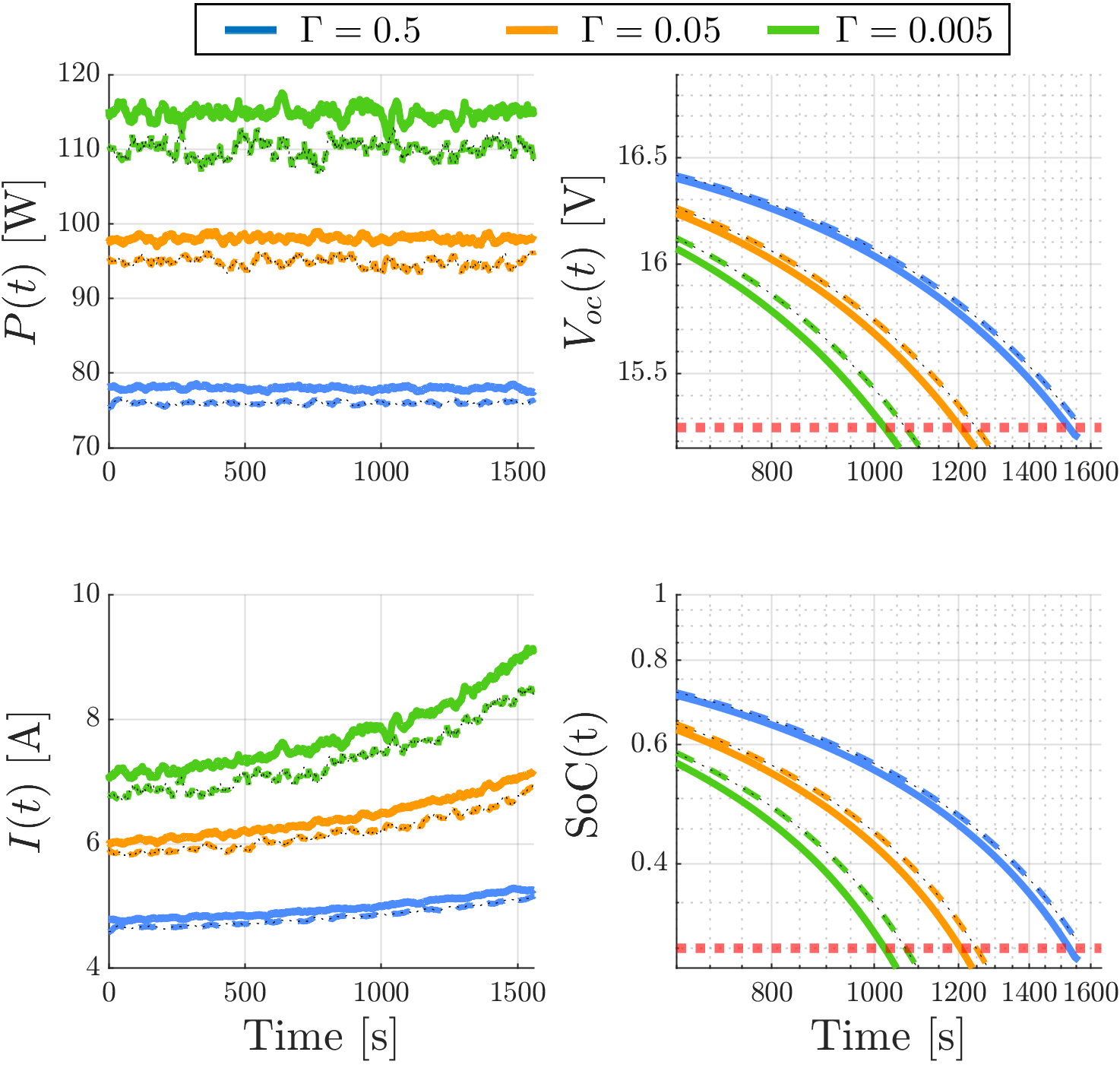}
\caption{Battery states: 1500-second (full-discharge) cycle.}
\label{fig:power_2}
\end{center}
\end{figure}
\\
Fig.~\ref{fig:power_2} expands the analysis to a 1500-second hover scenario, starting at a fully charged state—(SoC,$\, V_{\text{oc}}$)=(1.0,16.8$\, \mathrm{V}$)—and continuing until the safety threshold (dashed red line) is reached at (SoC,$\, V_{\text{oc}}$)=(0.3,15.26$\, \mathrm{V}$). While stabilizing against environmental noise, the LQG-regulated power consumption (top left) remains nearly constant throughout the flight. In contrast, the OCV (top right) exhibits a gradual decline as energy is depleted, with a semi-log scale used to illustrate the exponential nature of the voltage drop. Accordingly, more current is drawn (bottom left) to maintain steady power delivery, following $I(t) = P_{\text{hov}} / V(t)$. 
\\
Finally, the SoC trajectories (bottom right), reflecting energy percentage, resemble an endurance test—candidates farther right indicate greater energy efficiency via extended duration.
\begin{table}[b]
\centering
\caption{Power and current usage: Baseline vs C-ZUPT.}
\renewcommand{\arraystretch}{1.6}
\begin{tabular}{|c|c|c|c|c|c|c|} \hline
\multirow{2}{*}{\diagbox{Model}{$\Gamma$ \ }} & \multicolumn{3}{c|}{$P_{\text{avg}} \ [\mathrm{W}]$ \CC} & \multicolumn{3}{c|}{$I_{\text{avg}} \ [\mathrm{A}]$ \CC} \\ \cline{2-7}
 & 0.5  & 0.05 & 0.005  & 0.5  & 0.05 & 0.005 \\ \hline \hline 
Baseline \CC & 79.12 & 99.12 & 116.32 & 4.96 & 6.42 & 7.93 \\ \hline
Ours \CC & 75.74 & 93.66 & 108.02 & 4.78 & 6.13 & 7.43 \\ \hline \hline 
Reduction [\%] \CC & \textbf{4.27} & \textbf{5.51} & \textbf{7.14} & \textbf{3.63} & \textbf{4.52} & \textbf{6.31} \\ \hline 
\end{tabular} \label{t:err_5}
\end{table} 
Table~\ref{t:err_5} summarizes the full-operation results by comparing average power and current across varying $\Gamma$ intensities. 
\\
As expected, higher disturbance levels (lower $\Gamma$) consistently increase energy consumption; nevertheless, the C-ZUPT-aided variant (ours) achieves a 3\%–7\% reduction in these metrics.
Table~\ref{t:err_6} further reports the achieved flight times—defined at the dashed red safety threshold—on the left, alongside their corresponding efficiency on the right, expressed as flight minutes per unit of energy.
\begin{table}[h]
\centering
\caption{Estimated hover time: Baseline vs C-ZUPT.}
\renewcommand{\arraystretch}{1.6}
\begin{tabular}{|c|c|c|c|c|c|c|} \hline
\multirow{2}{*}{\diagbox{Model}{$\Gamma$ \ }} & \multicolumn{3}{c|}{$t_{\text{safe}}$ [min] \CC} & \multicolumn{3}{c|}{$\eta_{\text{eff}}$ [\text{min}/$\mathrm{W}$h] \CC} \\ \cline{2-7}
 & 0.5 & 0.05 & 0.005  & 0.5  & 0.05 & 0.005 \\ \hline \hline 
Baseline \CC & 25.5 & 20.1 & 17.0 & 0.573 & 0.452 & 0383 \\ \hline
Ours \CC & 26.2 & 20.9 & 18.1 & 0.591 & 0.472 & 0.407 \\ \hline \hline 
Improvement \CC & \textbf{0.77} & \textbf{0.88} & \textbf{1.07} & \textbf{2.9}\% & \textbf{4.2}\% & \textbf{5.9}\% \\ \hline 
\end{tabular} \label{t:err_6}
\end{table} 
\\
Evidently, C-ZUPT extends flight time by nearly one minute across all $\Gamma$, resulting in a 3\%-6\% improvement in overall system efficiency.
%

\section{Discussion and Conclusion} \label{sec:conc}
This work presents a sensor-free C-ZUPT state correction method for aerial navigation and control, independent of surface contact constraints. Based on a Newton–Euler framework, the C-ZUPT mechanism—triggered under near-stationary conditions—enables accurate state estimation, improved navigation, and control. Validation under varying disturbances shows consistent gains in hover stability, with notable reductions in estimation drift, control effort, and power consumption, alongside improved disturbance rejection.
The results show that the proposed C-ZUPT mechanism reduces control effort, extends flight time, and delays battery depletion to its safety threshold—thereby improving efficiency by lowering the power-to-weight ratio required to counteract gravity. These findings also underscore the method’s robust disturbance rejection, making it particularly well-suited for resource-constrained aerial platforms operating sensor-free at flexible update rates.
Nonetheless, several limitations warrant consideration:
\\
i) \textit{Point of no return}: C-ZUPT works best under mild dynamics as velocity errors remain bounded; however, once velocity estimates exceed a critical threshold, stable recovery without position fixes becomes unlikely, and triggering may fail.
\\
ii) \textit{Trade-off heuristics}: Tuning the window size and threshold values involves balancing sensitivity and robustness—values that are too small can lead to noisy false detections, while overly large values risk missing true events.
\\
iii) \textit{Context sensitivity}: Effective LQG control relies on carefully tuned gains and weights specific to each scenario. Thus, observed improvements apply to the tested platform, trajectories, and noise profiles, and may vary otherwise.
\\
Despite these limitations, the proposed approach offers a practical means of leveraging aerial stationarity intervals—a key asset in modern aviation. This enhanced performance can be decisive in time-critical hovering operations such as urban warfare, policing, rescue missions, surveillance, infrastructure inspection, environmental monitoring, and delivery services.
%


\section*{Acknowledgement}
D.E. is supported by the Maurice Hatter foundation and by the Bloom School Institutional Excellence Scholarship for outstanding doctoral students at the University of Haifa.

\section*{Appendix}
\renewcommand{\thesubsection}{\Alph{subsection}}

\subsection{Derivation of State Equations} \label{appendix:a}
This section presents an approximation of the nonlinear dynamics in \eqref{eq:x_dot} valid near the hovering equilibrium, where higher-order terms are omitted to facilitate linearization. 
\\
To begin with, let the following vectors denote the inertial positions, ${\boldsymbol{\xi}}^\mathcal{I}$, Euler angles ${\boldsymbol{\eta}}^\mathcal{I}$, and body rates, ${\boldsymbol{\omega}}^\mathcal{B}$, given by 
\begin{align} 
\boldsymbol{\xi}^\mathcal{I} = \begin{pmatrix} x \\ y \\ z \end{pmatrix} 
\ , \ 
\boldsymbol{\eta}^\mathcal{I} = \begin{pmatrix} \phi \\ \theta \\ \psi \end{pmatrix} 
\ , \ 
\boldsymbol{\omega}^\mathcal{B} = \begin{pmatrix} p \\ q \\ r \end{pmatrix} \ ,
\end{align} 
which together form the following state vector
\begin{align}
\boldsymbol{x} &= \begin{pmatrix} \,
({\boldsymbol{\xi}}^\mathcal{I})^\top & ({\boldsymbol{v}}^\mathcal{B})^\top & ({\boldsymbol{\eta}}^\mathcal{I})^\top & ({\boldsymbol{\omega}}^\mathcal{B})^\top \
\end{pmatrix}^\top \   \\
&= \begin{pmatrix} x & y & z  \ | \ u & v & w \ | \ \phi & \theta & \psi \ | \ p & q & r \end{pmatrix}^\top . \notag
\end{align} 
For stable horizontal hovering, the small-angle approximation ($s_\alpha \approx \alpha$ , $c_\alpha \approx 1$), simplifies the rotation matrix to
\begin{align}
\mathcal{R}_{\mathcal{B}}^{\mathcal{I}} \approx 
\begin{pmatrix}
1  & -\psi & \theta \\
\psi  &  1 & -\phi \\
-\theta  & \phi & 1
\end{pmatrix} = \boldsymbol{I}_3 + \boldsymbol{\eta} \times ,
\end{align}
for which, the inertial translations simplify to
\begin{align}
\dot{\boldsymbol{\xi}}^\mathcal{I} = \begin{pmatrix} \dot{x} \\ \dot{y} \\ \dot{z} \end{pmatrix} = 
\begin{pmatrix}
u - v \psi + \theta w \\
u \psi + v - \phi w \\
-u \theta + v \phi + w 
\end{pmatrix} = (\boldsymbol{I}_{3} + \boldsymbol{\eta} \times) \, {\boldsymbol{v}}^\mathcal{B} \, .
\end{align}
Similarly, the transformation of angular velocities from the body-frame to the inertial-frame simplifies to
\begin{align} 
\textbf{\textit{W}}_{\boldsymbol{\eta}} = 
\begin{pmatrix} 
1 & \text{s}_\phi \text{t}_\theta & \text{c}_\phi \text{t}_\theta \\
0 &  \text{c}_\phi & -\text{s}_\phi \\
0 & \text{s}_\phi / \text{c}_\theta & \text{c}_\phi / \text{c}_\theta 
\end{pmatrix} \approx \boldsymbol{I}_3 \, , 
\end{align} 
which leads to the simplified relation
\begin{align} 
\dot{\boldsymbol{\eta}}^\mathcal{I} = \textbf{\textit{W}}_{\boldsymbol{\eta}} {\boldsymbol{\omega}}^\mathcal{B} \approx {\boldsymbol{\omega}}^\mathcal{B} \, .
\end{align} 
The linear accelerations in the body frame are given by
\begin{align}
\dot{\boldsymbol{v}}^\mathcal{B} =& \ m^{-1} \boldsymbol{u}_z - {\boldsymbol{\omega}}^\mathcal{B} \times {\boldsymbol{v}}^\mathcal{B} - \textit{g} \, \textbf{e}^\mathcal{B}_z  \notag \\
=& \frac{1}{m} {u}_z \, \textbf{e}^\mathcal{B}_z - 
\begin{pmatrix}  
qw-rv \\ ru-pw \\ pv-qu
\end{pmatrix} 
-
\begin{pmatrix}  
- \text{s}_\theta \\
\text{s}_\phi \text{c}_\theta \\
\text{c}_\phi \text{c}_\theta 
\end{pmatrix} \textit{g} \, ,
\end{align}
however, under stable hovering conditions, small angular deviations allows us to neglect the cross-product terms, such that 
\begin{align}
\dot{\boldsymbol{v}}^\mathcal{B} = \begin{pmatrix}  
\dot{u} \\ \dot{v} \\ \dot{w} \end{pmatrix} = 
\begin{pmatrix}  
0 \\ 0 \\ 1/m
\end{pmatrix} {u}_z
-
\begin{pmatrix}  
- \theta \\ \phi \\ 1
\end{pmatrix} \textit{g} \, .
\end{align}
Lastly, the angular accelerations are given by
\begin{align}
\dot{\boldsymbol{\omega}}^\mathcal{B} = 
\begin{pmatrix} 
\dot{p} \\ \dot{q} \\ \dot{r}
\end{pmatrix} =
\begin{pmatrix} 
J_{xx}^{-1} \left( u_\phi + qr(J_{yy}-J_{zz}) \right) \\ 
J_{yy}^{-1} \left( u_\theta + pr(J_{zz}-J_{xx}) \right) \\ 
J_{zz}^{-1} \left( u_\psi + pq(J_{xx}-J_{yy}) \right) \\ 
\end{pmatrix} \, ,
\end{align}
however, as before, when the Coriolis terms are approximated to zero, the angular accelerations become
\begin{align}
\dot{\boldsymbol{\omega}}^\mathcal{B} = \boldsymbol{J}^{-1} \boldsymbol{\tau}^\mathcal{B}
\, .
\end{align}
Accordingly, to simplify the linearization process, the nonlinear system in \eqref{eq:x_dot} is approximated near the hover point as
\begin{align}
\begin{pmatrix}
\dot{\boldsymbol{\xi}}^\mathcal{I} \\ \dot{\boldsymbol{v}}^\mathcal{B} \\ \dot{\boldsymbol{\eta}}^\mathcal{I} \\ \dot{\boldsymbol{\omega}}^\mathcal{B} 
\end{pmatrix} = 
\begin{pmatrix}
( \boldsymbol{I}_{3} + \boldsymbol{\eta} \times ) \, {\boldsymbol{v}}^\mathcal{B} \\
m^{-1} \boldsymbol{u}_z - [ \textbf{e}_z^{\mathcal{I}} \times ] \, \textit{g} \\ 
%
\boldsymbol{\omega}^\mathcal{B} \\ 
\boldsymbol{J}^{-1} \left( \boldsymbol{u}_\phi + \boldsymbol{u}_\theta + \boldsymbol{u}_\psi  
\right)
\end{pmatrix} ,
\end{align}

\subsection{System Configuration Overview} \label{appendix:b}
Table~\ref{t:params} summarizes all parameters used in this study, grouped into four categories in the following top-to-bottom order:
\begin{enumerate}
    \item Rigid body mass properties~\cite{yu2020two}.
    \item Aerodynamic parameters~\cite{yu2020two}.
    \item Battery model~\cite{he2011evaluation}.
    \item KF and measurement noise statistics~\cite{engelsman2023data}.
\end{enumerate}
\begin{table}[h]
\centering
\caption{Quadrotor parameter nomenclature and units.}
\renewcommand{\arraystretch}{1.33}
\begin{tabular}{|c|l|c|c|}
\specialrule{1.1pt}{1pt}{1pt} 
\textbf{Parameter}\CC& \textbf{Physical property}\CC& \textbf{Value}\CC& \textbf{Units}\CC\\ \specialrule{1.1pt}{1pt}{1pt}
$J_{xx}$ & Roll moment of inertia & 0.0159 & kg$\cdot$m$^2$ \\ \hline
$J_{yy}$ & Pitch moment of inertia & 0.0140 & kg$\cdot$m$^2$ \\ \hline
$J_{zz}$ & Yaw moment of inertia & 0.0279 & kg$\cdot$m$^2$ \\ \hline
$l$ & Moment arm & 0.15 & m \\ \hline
$m$ & Mass & 0.9689 & kg \\ \hline \hline
$A_r$ & Rotor disk area & 0.0491 & m$^2$ \\ \hline
$C_M$ & Torque coefficient & 5.392$\times$10$^{-4}$ & - \\ \hline
$C_T$ & Thrust coefficient & 6.38$\times$10$^{-3}$ & - \\ \hline
$k_M$ & Torque constant & 6.33$\times$10$^{-8}$ & kg$\cdot$m$^2$/rad$^2$ \\ \hline
$k_T$ & Thrust constant & 6.01$\times$10$^{-6}$ & kg$\cdot$m/rad$^2$ \\ \hline
$\rho$ & Air density & 1.225 & kg/m$^3$ \\ \hline
$r$ & Blade radius & 0.125 & m \\ \hline
$\eta_{\text{rot}}$ & Rotor efficiency & 0.80 & - \\ \hline
$\tau_{\text{rot}}$ & Rotor time constant & 0.02 & s \\ \hline \hline
$C_{\text{bat}}$ & Battery capacity & 3,000 & mAh \\ \hline
$C_1$ & 1st-order RC capacitance & 2.5 & F \\ \hline
$R_1$ & 1st-order RC resistance & 0.05 & m$\Omega$ \\ \hline
$R_0$ & Instantaneous resistance & 0.04 & m$\Omega$ \\ \hline
$\boldsymbol{\nu}$ & OCV coefficients & [14, 4.8, -2] & V \\ \hline \hline
$\Delta t$ & Solution interval & 0.001 & s \\ \hline
$\sigma_{\eta}$ & Orientation noise & 0.001 & rad \\ \hline
$\sigma_{f}$ & Accel. noise density & 0.002 & m/s$^2$/$\sqrt{\text{Hz}}$ \\ \hline
$\sigma_{_{\textit{GNSS}}}$ & GNSS precision & $\approx$ 3.0 & m \\ \hline
$\sigma_{\textit{gyro}}$ & Gyro noise density & 0.001 & rad/s/$\sqrt{\text{Hz}}$ \\ \hline
$\sigma_{\omega}$ & Body rate noise & 0.001 & rad/s \\ \hline
$\sigma_{_{\textit{ZUPT}}}$ & ZUPT noise & 0.005 & m/s \\ 
\specialrule{1.1pt}{1pt}{1pt}
\end{tabular} \label{t:params}
\end{table}

\bibliographystyle{ieeetr}
\bibliography{Ref}

\end{document}